\documentclass[review]{elsarticle}
\usepackage{graphicx}

\usepackage{lineno,hyperref}
\usepackage{float}
\usepackage{comment}
\usepackage{multirow}
\usepackage{booktabs}
\modulolinenumbers[5]
\usepackage{xcolor}
\usepackage{subfigure}
\usepackage{svg}
\usepackage{soul}
\usepackage{tcolorbox}

\usepackage{amssymb}
\usepackage{amsmath}

\journal{Journal of \LaTeX\ Templates}

\begin{document}

\begin{frontmatter}
\title{Knowledge Graph informed Fake News Classification via Heterogeneous Representation Ensembles}
\tnotetext[mytitlenote]{Fully documented templates are available in the elsarticle package on \href{http://www.ctan.org/tex-archive/macros/latex/contrib/elsarticle}{CTAN}.}

\author{Boshko Koloski}
\address{Jo\v{z}ef Stefan Int. Postgraduate School}
\address{Jo\v{z}ef Stefan Institute\\1000 Ljubljana}

\author{Timen Stepi\v{s}nik Perdih}
\address{Jo\v{z}ef Stefan Institute\\1000 Ljubljana}

\author{Marko Robnik-\v{S}ikonja}
\address{University of Ljubljana, Faculty of Computer and Information Science \\ 1000 Ljubljana}

\author{Senja Pollak}
\address{Jo\v{z}ef Stefan Institute\\1000 Ljubljana}

\author{Bla\v{z} \v{S}krlj}
\address{Jo\v{z}ef Stefan Int. Postgraduate School}
\address{Jo\v{z}ef Stefan Institute\\1000 Ljubljana}




\begin{abstract}
\begin{tcolorbox}

 The final formatted version of this publication was published in Neurocomputing, Online, 29 January 2022 and is available at \url{https://doi.org/10.1016/j.neucom.2022.01.096}.

\end{tcolorbox}

Increasing amounts of freely available data both in textual and relational form offers exploration of richer document representations, potentially improving the model performance and robustness. An emerging problem in the modern era is fake news detection --- many easily available pieces of information are not necessarily factually correct, and can lead to wrong conclusions or are used for manipulation. In this work we explore how different document representations, ranging from simple symbolic bag-of-words, to contextual, neural language model-based ones can be used for efficient fake news identification. One of the key contributions is a set of novel document representation learning methods based solely on knowledge graphs, i.e. extensive collections of (grounded) subject-predicate-object triplets.
We demonstrate that knowledge graph-based representations already achieve competitive performance to conventionally accepted representation learners. Furthermore, when combined with existing, contextual representations, knowledge graph-based document representations can achieve state-of-the-art performance.
To our knowledge this is the first larger-scale evaluation of how knowledge graph-based representations can be systematically incorporated into the process of fake news classification.
\end{abstract}


\end{frontmatter}


\section{Introduction}
\label{sec:introduction}
Identifying fake news is a crucial task in the modern era. Fake news can have devastating implications on society; the uncontrolled spread of fake news can for example impact the idea of democracy, with the ability to alter the course of elections by targeted information spreading~\cite{allcott2017social}.
In the times of a global pandemic they can endanger the global health, for example by reporting that using bleach can stop the spread of Coronavirus~\cite{ijerph17072430, 10.1093/jtm/taaa057}, or that vaccines are problematic for human health. With the upbringings of the development of the information society, the increasing capability to create and spread news in various formats makes the detection of problematic news even harder. 

For media companies' reputation it is crucial to avoid distributing \textbf{unreliable information}. With the ever-increasing number of users and potential fake news spreaders, relying only on manual analysis is becoming unmanageable given the number of posts a single person can curate on a daily basis. Therefore, the need for \emph{automated detection} of fake news is more important than ever, making it also a very relevant and attractive research task. 

By being able to process large collections of labeled and unlabeled textual inputs, contemporary machine learning approaches are becoming a viable solution to automatic e.g., credibility detection{~\cite{shu2017fake}}. One of the key problems, however, concerns the representation of such data in a form, suitable for learning. Substantial advancements were made in this direction in the last years, ranging from large-scale curated knowledge graphs that are freely accessible to contextual language models capable of differentiating between subtle differences between a multitude of texts{~\cite{shu2020mining}. This work explores how such technologies can be used to aid and prevent spreading of problematic content, at scale.}
  
With the advancements in the field of machine learning and natural language processing, various different computer-understandable representations of texts have been proposed.  While the recent work has shown that leveraging background knowledge can improve document classification ~\cite{ostendorff2019enriching}, this path has not yet been sufficiently explored for fake news identification.
The main contributions of this work, which significantly extend our conference paper~\cite{koloski2021identification} are:

\begin{enumerate}
     \item We explore how additional background knowledge in the form of \textbf{knowledge graphs}, constructed from freely available knowledge bases can be exploited to enrich various contextual and non-contextual document representations.
    \item We conducted extensive
        experiments where we systematically studied the effect of five document and six different knowledge graph-based representations on the model performance.
    \item We propose a feature-ranking based \emph{post-hoc} analysis capable of pinpointing the key types of representation, relevant for a given classification problem.
    \item The explanations of the best-performing model are inspected and linked to the existing domain knowledge.
\end{enumerate}
    
\par The remaining work is structured as follows.
In Section~\ref{sec:related-work}, we present the relevant related work, followed by the text and graph representations used in our study in Section~\ref{sec:method}, we present the proposed method, followed by the evaluation in Section~\ref{sec:experiments}. We discuss the obtained results in Sections~\ref{sec:results-qa} and~\ref{sec:results-ql} and finish with the concluding remarks in Sections~\ref{sec:discussion} and~\ref{sec:conclusions}.
    
\section{Related Work}
\label{sec:related-work}
We next discuss the considered classification task and the existing body of literature related to identification/detection of fake news.
The fake news text classification task is defined as follows: given a text and a set of possible classes (e.g., fake and real) to which a text can belong, an algorithm is tasked with predicting the correct class label assigned to the text. Most frequently, fake news text classification refers to classification of data based on \textbf{social media}. The early proposed solutions to this problem used hand-crafted features of the authors (instances) such as
word and character frequencies ~\cite{potthast-etal-2018-stylometric}. Other fake news related tasks include the identification of a potential author as a spreader of fake news and the verification of facts. Many of the contemporary machine learning approaches are based on deep neural-network models~\cite{glazkova2020g2tmn}.

Despite the fact that the neural network based approaches outperform other approaches on many tasks, they are not directly \textbf{interpretable}. On the other side, more traditional machine learning methods such as symbolic and linear models are easier to interpret and reason with, despite being outperformed by contemporary deep-learning methods. To incorporate both viewpoints, a significant amount of research has been devoted to the field of \textbf{neuro-symbolic computing}, which aims to bring the robustness of neural networks and the interpretability of symbolic approaches together. For example, a recent approach explored document representation enrichment with symbolic knowledge (Wang et. al ~\cite{wang-etal-2014-knowledge}). In their approach, the authors tried enriching a two-part model: a text-based model consisting of statistical information about text and a knowledge model based on entities appearing in both the KG and the text. Further, Ostendorff et al. ~\cite{ostendorff2019enriching} explored a similar idea considering learning separate embeddings of knowledge graphs and texts, and later fusing them together into a single representation. An extension to the work of Ostendorff et al. was preformed by Koloski et al. ~\cite{KOLOSKI_2020}, where a promising improvement of the joint representations has been observed. This approach showed potentially useful results, improving the performance over solely text-based models. 

\par Versatile approaches achieve state of the art results when considering various tasks related to fake news detection; Currently, the transformer architecture ~\cite{vaswani2017attention} is commonly adopted for various down-stream learning tasks. The winning solution to the COVID-19 Fake News Detection task ~\cite{patwa2020fighting} utilized fine-tuned BERT model that considered Twitter data scraped from the COVID-19 period - January 12 to
April 16, 2020~\cite{muller2020covid, glazkova2020g2tmn}. Other solutions exploited the recent advancements in the field of Graph Neural Networks and their applications in these classification tasks~\cite{gnn_fake_news}. However, for some tasks best preforming models are SVM-based models that consider more traditional n-gram-based representations~\cite{buda:2020}. 
Interestingly, the stylometry based approaches were shown~\cite{schuster-etal-2020-limitations} to be a potential threat for the automatic detection of fake news.
The reason for this is that machines are able to generate consistent writings regardless of the topic, while humans tend to be biased and make some inconsistent errors while writing different topics. Additionally researchers explored how the traditional machine learning algorithms perform on such tasks given a single representation \cite{gilda}. The popularity of deep learning and the successes of Convolutional and Recurrent Neural Networks motivated development of models following these architectures for the tasks of headline and text matching of an article \cite{ullah_cnn}. Lu and Li \cite{lu2020gcan} proposed a solution to a more realistic scenario for detecting fake news on social media platforms which incorporated the use of graph co-attention networks on the information about the news, but also about the authors and spread of the news.
However, individual representations of documents suitable for solving a given problem are mostly problem-dependent, motivating us to explore \emph{representation ensembles}, which potentially entail different aspects of the represented text, and thus generalize better.


\section{Proposed methodology}
\label{sec:method}
\begin{figure}[t!]
    \centering
    \includegraphics[width=\linewidth]{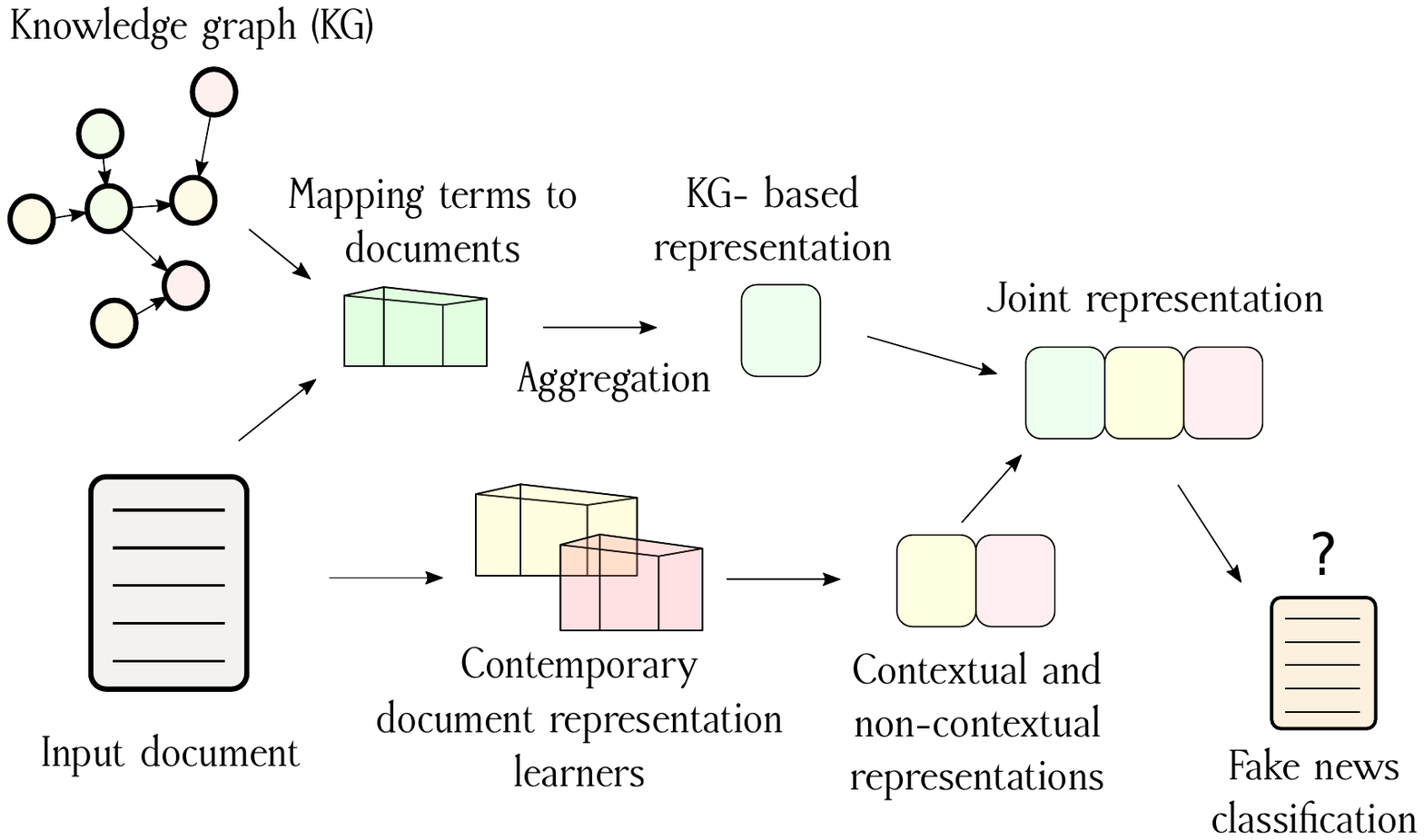}
    \vspace{-1.5cm}
    \caption{Schematic overview of the proposed methodology. Both knowledge graph-based features and contextual and non-contextual document features are constructed, and used simultaneously for the task of text classification.}
    \label{fig:fig_schema_method}
\end{figure}

In this section we explain the proposed knowledge-based representation enrichment method. First we define the relevant document representations, followed by concept extraction and knowledge graph (KG) embedding. Finally, we present the proposed combination of the constructed feature spaces.
Schematic overview of the proposed methodology is shown in Figure~\ref{fig:fig_schema_method}.
We begin by describing the bottom part of the scheme (yellow and red boxes), followed by the discussion of KG-based representations (green box). Finally, we discuss how the representations are combined ("Joint representation") and learned from (final step of the scheme).
\subsection{Existing document representations considered}    
\label{subsec:text_representaitons}
Various document representations capture different patterns across the documents. For the text-based representations we focused on exploring and exploiting the methods we already developed in our submission to the COVID-19 fake news detection task \cite{koloski2021identification}. We next discuss the document representations considered in this work.
\begin{description}
\item \textbf{Hand crafted features}.
    We use stylometric features inspired by early work in authorship attribution \cite{potthast-etal-2018-stylometric}. We focused on word-level and character-level statistical features.
\item \textbf{Word based features}. The word based features included maximum and minimum word length in a document, average word length, standard deviation of the word length in document. Additionally we counted the number of words beginning with upper and the number of words beginning a lower case.
\item \textbf{Character based features} The character based features consisted of the counts of digits, letters, spaces, punctuation, hashtags and each vowel, respectively.
Hence, the final statistical representation has 10 features.

\item \textbf{Latent Semantic Analysis}.
\label{subsub:lsa}
    Similarly to Koloski et al. \cite{koloski2020multilingual} solution to the PAN 2020 shared task on Profiling Fake News Spreaders on Twitter \cite{fnspreaders2020}
    we applied the low dimensional  space estimation technique. First, we preprocessed the data by lower-casing the document content and removing the hashtags, punctuation and stop words.
    From the cleaned text, we generated the POS-tags using the NLTK library\cite{nltk}. Next, we used the prepared data for feature construction.  For the feature construction we used the technique used by Martinc et al. \cite{martincPan} which iteratively weights and chooses the best n-grams.
    We used two types of n-grams: Word based: n-grams of size 1 and 2 and Character based: n-grams of sizes 1, 2 and 3.
   We generated word and character n-grams and used TF-IDF for their weighting. We performed SVD \cite{halko2009finding} of the TF-IDF matrix, where we only selected the $m$ most-frequent n-grams from word and character n-grams. With the last step we obtained the LSA representation of the documents. For each of our tasks, our final representation consists of  $2{,}500$ word and $2{,}500$ character features (i.e. $5{,}000$ features in total) reduced to  $512$ dimensions with the SVD.

\item \textbf{Contextual features}.
For capturing contextual features we utilize embedding methods that rely on the transformer architecture \cite{vaswani2017attention}, including: 
    \begin{itemize}
        \item DistilBert \cite{distilbert}  \textit{distilbert-base-nli-mean-tokens} - d = 768 dimensions 
        \item RoBERTa \cite{roberta} -  \textit{roberta-large-nli-stsb-mean-tokens} - d = 768 dimensions 
        \item XLM  \cite{xlm} - \textit{xlm-r-large-en-ko-nli-ststb} - d = 768 dimensions 
    \end{itemize}
    First, we applied the same preprocessing as described in subsection \ref{subsub:lsa}.
    After we obtained the preprocessed texts we embedded every text with a given transformer model and obtained the contextual vector representation. As the transformer models work with a limited number of tokens, the obtained representations were 512-dimensional, as this was the property of the used pre-trained models. This did not represent a drawback since most of the data available was shorter than this maximum length. The contextual representations were obtained via pooling-based aggregation of intermediary layers~\cite{reimers-2019-sentence-bert}.
 \end{description}
 
\subsection{Knowledge graph-based document representations}
\label{subsec:kg_representations}
We continue the discussion by presenting the key novelty of this work: document representations based solely on the existing background knowledge.
To be easily accessible, human knowledge can be stored as a collection of facts in knowledge bases (KB). 
The most common way of representing human knowledge is by connecting two entities with a given relationship that relates them. Formally, a knowledge graph can be understood as a directed multigraph, where both nodes and links (relations) are typed. A concept can be an abstract idea such as a thought, a real-world entity such as a person e.g., Donald Trump, or an object - a vaccine, and so on. An example fact is the following: Ljubljana (entity) is the capital(relation) of Slovenia(entity), the factual representation of it is \textit{(Ljubljana,capital,Slovenia)}. Relations have various properties, for example the relation \textit{sibling} that captures the symmetry-property - if (Ann,siblingOf,Bob) then (Bob,siblingOf,Ann), or antisymmetric relation fatherOf (Bob,fatherOf,John) then the reverse does not hold (John,fatherOf,Bob). 

 In order to learn and extract patterns from facts the computers need to represent them in useful manner. 
 To obtain the representations we use six knowledge graph embedding techniques: TransE \cite{transE}, RotatE\cite{sun2019rotate}, QuatE\cite{quatE}, ComplEx\cite{complEx}, DistMult\cite{distmult} and SimplE\cite{simplE}. The goal of a knowledge graph embedding method is to obtain numerical representation of the KG, or in the case of this work, its entities.  The considered KG embedding methods also aim to preserve relationships between entities. The aforementioned methods and the corresponding relationships they preserve are listed in Table \ref{tab:kg_rel}. It can be observed that RotatE is the only method capable of modeling all five relations.
    \begin{table}[htb!]
            \centering
               \caption{Relations captured by specific knowledge graph embedding from the GraphVite knowledge graph suite \cite{graphVite}. }
               \vspace{0.3cm}
            \resizebox{\textwidth}{!}{\begin{tabular}{|c|c|c|c|c|c|}
                \hline
                Name & Symmetry & Anti-symmetry & Inversion & Transitivity & Composition  \\ \hline
                TransE \cite{transE} & \textbf{x} & \textbf{x} & \checkmark  & \checkmark  & \textbf{x} \\ \hline
                DistMult \cite{distmult} & \checkmark & \textbf{x} & \textbf{x}  & \textbf{x} & \textbf{x} \\ \hline
                ComplEx  \cite{complEx} & \checkmark & \checkmark & \checkmark  & \checkmark & \textbf{x} \\ \hline
                RotatE \cite{sun2019rotate} & \checkmark & \checkmark & \checkmark  & \checkmark & \checkmark \\ \hline
                QuatE \cite{quatE} & \checkmark & \checkmark & \checkmark  & \checkmark & \textbf{x} \\ \hline
                SimplE \cite{simplE} & \checkmark & \checkmark & \checkmark  & \checkmark & \textbf{x} \\ \hline
            \end{tabular}}
            \label{tab:kg_rel}
        \end{table} 

Even though other methods are theoretically not as expressive, this does not indicate their uselessness when considering construction of document representations. For example, if transitivity is crucial for a given data set, and two methods, which theoretically both model this relation capture it to a different extent, even simpler (and faster) methods such as TransE can perform well.
We propose a novel method for combining background knowledge in the form of a knowledge graph \textit{KG} about concepts \textit{C} appearing in the data \textit{D}. To transform the documents in numerical spaces we utilize the techniques described previously. For each technique we learn the space separately and later combine them in order to obtain the higher dimensional spaces useful for solving a given classification task.

\begin{figure}[H]
    \centering
    \includegraphics[width=.7\linewidth]{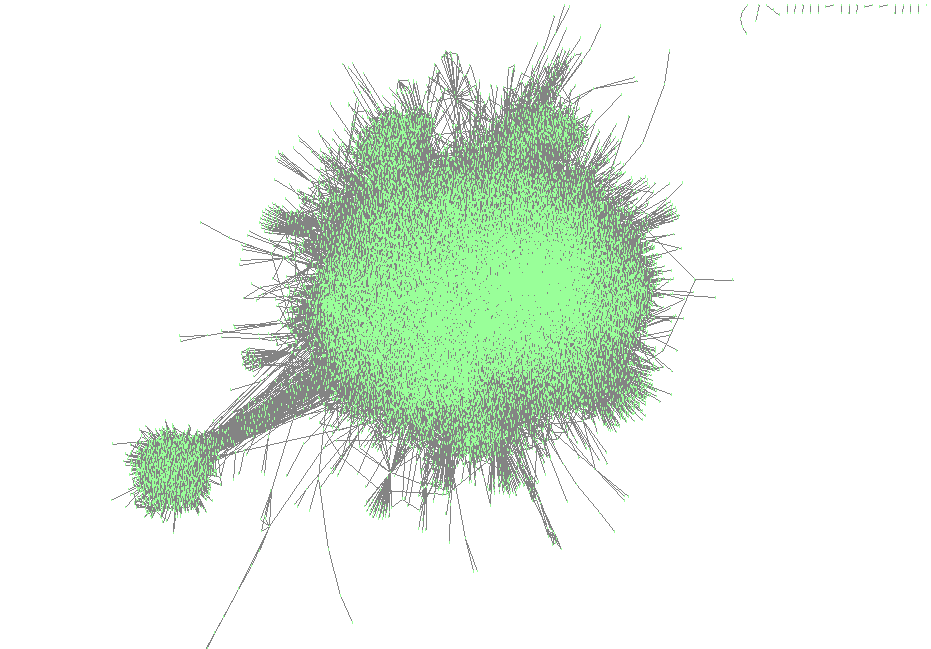}
    \caption{The WikiData5m knowledge graph - the $\approx$100{,}000 most connected nodes. It can be observed that multiple smaller structures co-exist as part of the global, well connected structure.}
    \label{fig:wng}
\end{figure}

For representing a given document, the proposed approach can consider the document text or also account for additional metadata provided for the document (e.g. the author of the text, their affiliation, who is the document talking about etc.). In the first case,  we identify which concept embeddings map to a given piece of text, while in the second scenario we also embed the available metadata and jointly construct the final representation.  In this study we use the WikiData5m knowledge graph \cite{wikidata} (Figure~\ref{fig:wng}).  The most central nodes include terms such as `encyclopedia' and `united state'.

The GraphVite library \cite{graphVite} incorporates approaches that map aliases of concepts and entities into their corresponding embeddings. To extract the concepts from the documents we first preprocess the documents with the following pipeline:
punctuation removal; stopword removal for words appearing in the NLTK's english stopword list; lemmatization via the \textit{NLTK's WordNetLemmatizer} tool.
 
 In the obtained texts, we search for concepts (token sets) consisting of uni-grams, bi-grams and tri-grams, appearing in the knowledge graph. 
 The concepts are identified via exact string alignment.
 With this step we obtained a collection of candidate concepts $C_{d}$ for each document $d$.

From the obtained candidate concepts that map to each document, we developed three different strategies for constructing the final representation. Let $\boldsymbol{e}^i$ represent the $i$-th dimension of the embedding of a given concept. Let $\bigoplus$ represent the element wise summation ($i$-th dimensions are summed). We consider the following aggregation. We considered using all the concepts with equal weights and obtained final concept as the average of the concept embeddings:
        \begin{equation*}
            \textsc{agg-average}(C_d) = \frac{1}{|C_d|}\bigoplus_{c \in C_d} \boldsymbol{e}_c.
        \end{equation*}
The considered aggregation scheme, albeit being one of the simpler ones, already offered document representations competitive to many existing mainstream approaches. The key parameter for such representations was embedding dimension, which was in this work set to 512.

\subsection{Construction of the final representation}
Having presented how document representations can be obtained from knowledge graphs, we next present an overview of the considered document representations used for subsequent learning, followed by the considered representation combinations. The overview is given in Table~\ref{tab:representation_tab}.
\begin{table}[htb!]
    \centering
       \caption{Summary table of the textual and KG representations used in this paper.}
       \vspace{0.3cm}
    \resizebox{\textwidth}{!}{\begin{tabular}{|c|c|c|c|}
        \hline
        Name & Type & Description & Dimension  \\ \hline
        Stylomteric & text & Statistical features capturing style of an author. & 10 \\
        LSA & text & N-gram based representations built on chars and words reduced to lower dimension via SVD. & 512 \\
        DistilBert & text & Contextual  - transformer based representation learned via sentence-transformers. & 768 \\
        XLM & text & Contextual - transformer based representation learned via sentence-transformers. &768  \\
        RoBERTa & text& Contextual -  - transformer based representation learned via sentence-transformers. &  768  \\
        \hline
        TransE & KG & KG embedding capturing inversion, transitivity and composition property. &  512 \\
        DistMult & KG & KG embedding capturing symmetry property. & 512 \\
        ComplEx & KG & KG embedding capturing symmetry, anti-symmetry, inversion and transitivity property. & 512 \\
        RotatE & KG & KG embedding captures inversion, transitivity and composition property. & 512 \\
        QuatE & KG & KG embedding capturing symmetry, anti-symmetry, inversion, transitivity and composition property. & 512 \\
        SimplE & KG & KG embedding capturing symmetry, anti-symmetry, inversion and transitivity property. & 512 \\
        \hline
    \end{tabular}
}
    \label{tab:representation_tab}
\end{table}
Overall, 11 different document representations were considered. Six of them are based on knowledge graph-based embedding methods. The remaining methods either consider contextual document representations (RoBERTa, XLM, DistilBert), or non-contextual representations (LSA and stylometric). The considered representations entail multiple different sources of relevant information, spanning from single character-based features to the background knowledge-based ones. 

For exploiting the potential of the multi-modal representations we consider three different scenarios to compare and study the potential of the representations:   
\begin{description}
    \item \textbf{LM} - we concatenate the representations from Section \ref{subsec:text_representaitons} - handcrafted statistical features, Latent Semantic Analysis features, and contextual representations - XLM, RoBERTa and DistilBERT.
    \item \textbf{KG} - we concatenate the aggregated concept embeddings for each KG embedding method from Subsection \ref{subsec:kg_representations} - TransE TransE, SimplE, ComplEx, QuatE, RotatE and DistMult. We agreggate the concepts with the \textsc{AGG-AVERAGE} strategy.
    \item \textbf{Merged} - we concatenate the obtained language-model and knowledge graph representations. As previously mentioned we encounter two different scenarios for KG: 
    \begin{itemize}
        \item LM+KG - we combine the induced KG representations with the methods explained in Subsection \ref{subsec:kg_representations}.
        \item LM+KG+KG-ENTITY - we combine the document representations, induced KG  representations from the KG and the metadata KG representation if it is available. 
       To better understand how the metadata are used (if present), consider the following example. Consider a document, for the author of which we know also the following information:
       \textit{speaker = Dwayne Bohac, job = State representative, subject = abortion, country = Texas , party affiliation = republican}. The values of such metadata fields (e.g., job) are considered as any other token, and checked for their presence in the collection of knowledge graph-based entity embeddings. Should the token have a corresponding embedding, it is considered for constructing the KG-ENTITY representation of a given document. For the data sets where the metadata is present, it is present for all instances (documents). If there is no mapping between a given collection of metadata and the set of entity embeddings, empty (zero-only) representation is considered.
\end{itemize}
\end{description}
Having discussed how the constructed document representation can be combined systematically, we next present the final part needed for classification -- the representation ensemble model construction.
\subsection{Classification models considered}
\label{sec:stacking}
We next present the different neural and non-neural learners, which consider the constructed representations discussed in the previous section.

\textbf{Representation stacking with linear models}.
The first approach to utilize the obtained representations was via linear models that took the stacked representations and learned a classifier on them. We considered using a LogisticRegression learner and a StochasticGradientDescent based learner that were optimized via either a \textit{log} or \textit{hinge} loss function. We applied the learners on the three different representations scenarios.
    
\begin{figure}[H]
        \centering
        \resizebox{\textwidth}{!}{\includegraphics{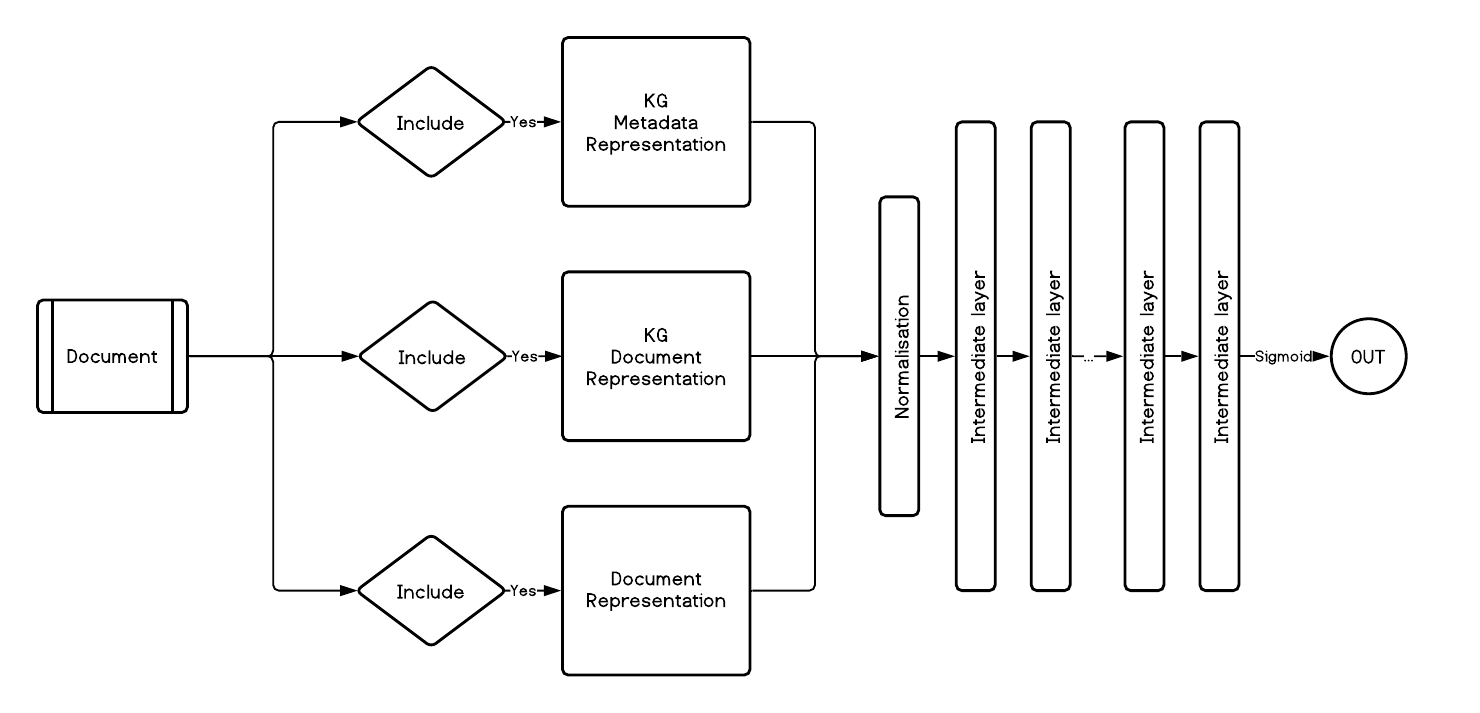}}
        \caption{Neural network architecture for learning the joint intermediate representations. The \textit{Include} decision block implies that some of the representations can be optionally excluded from the learning. The number of the intermediate layers and the dimensions are of varying sizes and are part of the model's input.}
        \label{fig:nn_schema}
    \end{figure}

\textbf{Representation stacking with neural networks}.
    Since we have various representations both for the textual patterns and for the embeddings of the concepts appearing in the data we propose an intermediate joint representation to be learnt with a neural network. For this purpose, we propose stacking the inputs in a heterogeneous representation and learning intermediate representations from them with a neural network architecture. The schema of our proposed neural network approach is represented in Figure \ref{fig:nn_schema}.
    We tested three different neural networks for learning this task. \\
     The proposed architecture consists of main two blocks: the input block and the hidden layers-containing block. The input block takes the various representations as parameters and produces a single concatenated representation which is normalized later. The hidden layer block is the learnable part of the architecture, the input to this block are the normalized representations and the number of the intermediate layers as well as their dimension. We evaluate three variants of the aforementioned architecture:
     \begin{description}
    \item \textbf{[SNN] Shallow neural network}. In this neural network we use a single hidden layer to learn the joint representation. 
    \item \textbf{[5Net] Five hidden layer neural network}. The original approach that we proposed to solve the COVID-19 Fake News Detection problem featured a five layer neural network to learn the intermediate representation \cite{koloski2021identification}. We alter the original network with the KG representations for the input layer.
    \item \textbf{[LNN] Log(2) scaled neural network}. Deeper neural networks in some cases appear to be more suitable for some representation learning tasks. To exploit this hypothesis we propose a deeper neural network - with a domino based decay. For $n$ intermediate layers we propose the first intermediate layer to consist of $2^{n}$ neurons, the second to be with $2^{n-1}$ ... and the $n_0$-th to be activation layer with the number of unique outputs. 

\end{description}
\section{Empirical evaluation}
\label{sec:experiments}
In this section, we first describe four data sets which we use for benchmarking of our method. Next we discuss the empirical evaluation of the proposed method, focusing on the problem of fake news detection.
\subsection{Data sets}
    In order to evaluate our method we use four different fake news problems. We consider a fake news spreaders identification problem, two binary fake news detection problems and a multilabel fake news detection problem. We next discuss the data sets related to each problem considered in this work. 
    \begin{description}
        \item  \textit{COVID-19 Fake News} detection data set \cite{patwa2020fighting,patwa2021overview} is a collection of social media posts from various social media platforms Twitter, Facebook, and YouTube. The data contains COVID-19 related posts, comments and news, labeled as \textit{real} or \textit{fake}, depending on their truthfulness.  Originally the data is split in three different sets: train, validation and test. 
        \item  \textit{Liar, Liar Pants on Fire} \cite{wang-2017-liar} represents a subset of PolitiFact's collection of news that are labeled in different categories based on their truthfulness. PolitiFact represents a fact verification organization that collects and rates the truthfulness of claims by officials and organizations. This problem is multi-label classification based with six different degrees of a fake news provided. For each news article, an additional metadata is provided consisting of: speaker, controversial statement, US party to which the subject belongs, what the text address and the occupation of the subject.
        \item    \textit{Profiling fake news Spreaders} is an author profiling task that was organized under the PAN2020 workshop \cite{fnspreaders2020}. In author profiling tasks, the goal is to decide if an author is a spreader of fake news or not, based on a collection of posts the author published. The problem is proposed in two languages English and Spanish. For each author 100 tweets are given, which we concatenate as a single document representing that author. 
        \item \textit{FNID: FakeNewsNet} \cite{fnid_fnn} is a data set containing news from the PolitiFact website. The task is binary classification with two different labels - real     and fake. For each news article - fulltext, speaker and the controversial statement are given.
    \end{description}
    
The data splits are summarised in Table \ref{tab:all_datsets}.
\begin{table}[htb!]
    \centering
    \begin{tabular}{|r|r|r|r|r|}
     \hline
     data set & Label & Train & Validation & Test \\ \hline
     \multirow{3}{*}{\rotatebox[origin=c]{0}{COVID-19}}  & real & 3360 (52\%) & 1120 (52\%) & 1120 (52\%)   \\
     & fake & 3060 (48\%) & 1020 (48\%) & 1020 (48\%) \\ \cline{2-5}
      & all & 6420 (100\%) & 2140 (100\%) & 2140 (100\%) \\  \hline
     \multirow{3}{*}{\rotatebox[origin=c]{0}{{PAN2020}}} & real & 135 (50\%) & 15 (50\%)  & 100 (50\%) \\ 
      & fake & 135 (50\%) & 15 (50\%)  & 100 (50\%) \\ \cline{2-5}
      & all & 270 (100\%) & 30 (100\%) & 200 (100\%) \\ \hline
    \multirow{3}{*}{\rotatebox[origin=c]{0}{FakeNewsNet}} & real & 7591 (50.09\%) & 540 (51.03\%) & 1120 (60.34\%)   \\ 
      & fake & 7621 (49.91\%) & 518 (48.96\%) & 1020 (39.66\%) \\ \cline{2-5}
      & all & 15212 (100\%) & 1058 (100\%) & 1054 (100\%)\\ \hline
     \multirow{7}{*}{\rotatebox[origin=c]{0}{LIAR}} & barely-true & 1654 (16.15\%) & 237 (18.46\%) & 212 (16.73\%) \\ 
      & false & 1995 (19.48\%) & 263 (20.48\%) & 249 (19.65\%)   \\
      & half-true & 2114 (20.64\%) & 248 (19.31\%) & 265 (20.92\%) \\ 
      & mostly-true & 1962 (19.16\%) & 251 (19.55\%) & 241 (19.02\%) \\ 
      & pants-fire & 839 (8.19\%) & 116 (9.03\%) & 92 (7.26\%) \\ 
      & true & 1676 (16.37\%) & 169 (13.16\%) & 208 (16.42\%)   \\ \cline{2-5}
      & all & 10240 (100\%) & 1284 (100\%) & 1267 (100\%)\\ \hline 
    \end{tabular}
    \caption{Distribution of samples per given label in the three splits: train, validation and test for all four data sets respectively.}
    \label{tab:all_datsets}
\end{table}

\subsection{Document to knowledge graph mapping}
 For each article we extract the uni-grams, bi-grams and tri-grams that also appear in the Wikidata5M KG. Additionally, for the \textit{Liar} and the \textit{FakeNewsNet} data sets we provided KG embedding based on the aggregated concept embedding from their metadata. In the case of the \textit{Liar} data set we use if present the speaker, the party he represents, the country the speech is related with and the topic of their claim.
 In all evaluation experiments we use the \textsc{AGG-AVERAGE} aggregation of concepts.
\subsection{Classification setting}
We use the train splits of each data set to learn the models, and use the validation data splits to select the best-performing model to be used for final test set evaluation. For both the linear stacking and the neural stacking we define custom grids for hyperparameter optimization, explained in the following subsections.
    
    \par \textbf{Learning of linear models}
    For each problem we first learn a baseline model from the given representation and a L2 regularized Linear Regression with the parameter $\lambda_{2} \in \{0.1,0.01,0.001\}$. We also learned StochasticGradientDescent(SGD)-based linear learner optimizing 'log' and 'hinge' functions with ElasticNet regularization. For the SGD learner we defined a custom hyperparameter grid:
    \begin{description}
    \item   $l1\_ratio \in \{0.05,0.25,0.3,0.6,0.8,0.95\}$,
    \item $power\_t \in \{0.1,0.5,0.9\}$,
    \item $alpha \in \{0.01,0.001,0.0001,0.0005\}$.
    \end{description}
      
    \par \textbf{Learning of neural models}
    The optimization function for all of the neural models was the CrossEntropyLoss optimized with the Adam Optimizer \cite{adam}. We used the \textit{SELU} function as an activation function between the intermediate layers. For fine-tuning purposes we defined a custom grid consisting of the learning rate $\lambda$, the dropout rate $p$ and the number of intermediate layers $n$ (for each network separately). The search-spaces of each parameter are:
    \begin{description}
        \item Learning rate: $\lambda \in \{0.0001, 0.005, 0.001, 0.005, 0.01, 0.05, 0.1\}$.
        \item Dropout rate: $p \in \{0.1, 0.2, 0.3, 0.4, 0.5, 0.6, 0.7, 0.8, 0.9\}$.
        \item Intermediate layer parameters:
        \begin{itemize}
            \item SN $n \in \{32,64,128,256,512,1024,2048,4096,8192,16384\}$.
            \item 5Net fixed sizes as in \cite{koloski2021identification}.
            \item LNN $n \in {6,7,8,9,10,11,12,13,14,15,16}$ which produced $n$. intermediate layers of sizes $2^{n}, 2^{n-1}, 2^{n-2},...,2^{2},2$. Note that in total, ten different architectures were tested.
        \end{itemize}
    \end{description}
We considered batches of size $32$, and trained the model for a maximum of $1{,}000$ epochs with an early stopping criterion - if the result did not improve for $10$ successive epochs we stopped the optimization.

\subsection{Baselines}
\label{sec:baselines}
The proposed representation-learner combinations were trained and validated by using the same split structure as provided in a given shared task, hence we compared our approach to the state-of-the-art for each data set separately. As the performance metrics differ from data set to data set, we compare our approach with the state-of-the-art with regard to the metric that was selected by the shared task organizers.

\section{Quantitative results}
\label{sec:results-qa}

In this section, we evaluate and compare the quality of the representations obtained for each problem described in Section \ref{sec:experiments}. For each task we report four metrics: \textit{accuracy}, \textit{F1-score}, \textit{precision} and \textit{recall}.

\subsection{Task 1: LIAR} The best-performing model on the validation set was a \textbf{[SNN]} shallow neural network with 128 neurons in the intermediate layer, a learning rate of 0.0003, batch size of 32, and a dropout rate of 0.2. The combination of the textual and KG representations improved significantly over the baseline models. The best-performing representations were constructed from the language model and the KG entities including the ones extracted from the metadata. The assembling of representations gradually improves the scores, with the combined representation being the top performing our model. The metadata-entity based representation outperforms the induced representations by a margin of $2.42\%$, this is due the captured relations between the entities from the metadata. The evaluation of the data is task with respect to the models is shown in Table \ref{tab:eval_LIAR}.
\begin{table}[H]
        \centering
        \caption{Comparison of representations on the \textit{Liar} data set without background knowledge (LM) with models incorporating text knowledge graph embeddings (KG) and metadata knowledge graph-embeddings (KG-ENTITY). LR in the representation column denotes the linear regression learner and SNN indicates the shallow neural network. The introduction of the factual knowledge continually improved the performance of the model.}\vspace{0.3cm}
        \resizebox{\textwidth}{!}{\begin{tabular}{|l|c|c|c|c|c|}
            \hline
            Representation & Accuracy & F1 - score & Precision & Recall  \\ \hline
            LR(LM) &  0.2352 & 0.2356 &  0.2364 &  0.2352 \\ \hline
            LR(KG) &  0.1996 & 0.1993 &  0.2004 &  0.1997 \\ \hline
            LR(LM + KG) &  0.2384 & 0.2383 & 0.2383 &  0.2384\\ \hline
            LR(KG-ENTITY) &   0.2238 & 0.2383 & 0.2418 &  0.2415 \\ \hline \hline
            LR(LM + KG-ENTITY) & 0.2399 &  0.2402 & 0.2409 & 0.2399 \\  \hline
            LR(LM + KG + KG-ENTITY) & 0.2333 & 0.2336 & 0.2332 & 0.2336 \\  \hline
            SNN(LM + KG + KG-ENTITY) & 0.2675 &  \textbf{0.2672} & \textbf{0.2673} & \textbf{0.2676} \\ \hline \hline
            SOTA (literature)  \cite{alhindi-etal-2018-evidence} & \textbf{0.3740} & x & x & x \\ \hline
          \end{tabular}}
        \label{tab:eval_LIAR}
\end{table}

\subsection{Task 2: FakeNewsNet} The Log(2) neural network was the best performing one for the \textit{FakeNewsNet} problem with the n-parameter set to 12, a learning rate of 0.001, and a dropout rate of 0.7. The constructed KG representations outperformed both the LM representation by 1.99\% and the KG-ENTITY representation by 2.19\% in terms of accuracy and also outperformed them in terms of F1-score. The further combination of the metadata and the constructed KG features introduced significant improvement both with the linear stacking and the joint neural stacking, improving the baseline score by $1.23\%$ for accuracy, $1.87\%$ for F1-score and $3.31\%$ recall for the linear stacking. The intermediate representations outscored every other representation by introducing $12.99\%$ accuracy improvement, $13.32\%$ improvement of F1-score and $26.70\%$ gain in recall score. The proposed methodology improves the score over the current best performing model by a margin of $3.22\%$. The evaluation of the data is task with respect to the models is shown in Table \ref{tab:eval_FNN}.

\begin{table}[htb!]
        \centering
        \caption{Comparison of representations on the \textit{FakeNewsNet} data set without background knowledge (LM) with models incorporating text knowledge graph embeddings (KG) and metadata knowledge graph-embeddings (KG-ENTITY). LR in the representation column denotes the linear regression learner and LNN indicates the use of the Log(2) neural network.}\vspace{0.3cm}
        \resizebox{\textwidth}{!}{\begin{tabular}{|l|c|c|c|c|c|}
            \hline
            Representation & Accuracy & F1 - score & Precision & Recall  \\ \hline
            LR(LM) & 0.7581 & 0.7560 & 0.9657 & 0.6210  \\  \hline
            LR(KG) &  0.7780 & 0.7767 & \textbf{0.9879} & 0.6399 \\  \hline
            LR(LM+KG) & 0.7676 & 0.7704 & 0.9536 & 0.6462 \\  \hline
            LR(KG-ENTITY) & 0.7561 & 0.7512 & 0.9773 & 0.6100  \\  \hline \hline
            LR(LM + KG-ENTITY) & 0.7600 &  0.7602 & 0.9570 & 0.6305 \\  \hline
            LR(LM + KG + KG-ENTITY) & 0.7704 &  0.7747 & 0.9498 & 0.6541 \\  \hline
            LNN(LM + KG + KG-ENTITY) & \textbf{0.8880} &  \textbf{0.8892} & 0.9011 & \textbf{0.8880} \\ \hline \hline
            SOTA (literature) \cite{bidgoly2020fake} & 0.8558 & x & x & x \\ 
            \hline
          \end{tabular}}
        \label{tab:eval_FNN}
\end{table}
\subsection{Task 3: PAN2020}
For the \textit{PAN2020} problem, the best performing model uses the combination of the LSA document representation and the TransE and RotatE document representations and SGD based linear model on the subsets of all of the representations learned. The deeper neural networks failed to learn the intermediate representations more successfully due to the lack of data examples(only 300 were provided). The addition of factual knowledge (embedded with the TransE and RotatE methods) to the text representation improved the score of the model improving the LM based representation by 10\% gain in accuracy, and 8.59\% gain in F1-score. 

For the  \textit{PAN2020} problem, the best performing model uses the combination of the LSA document representation and the TransE and RotatE document representations and SGD based linear model on the subsets of all of the representations learned. The deeper neural networks failed to exploit the intermediate representations to a greater extent due to the lack of data examples(only 300 examples provided for the training). However, the problem benefited increase in performance with the introduction of KG-backed representations, gaining $5.5\%$ absolute improvement over the LM-only representation. The low amount of data available for training made the neural representations fail behind the subset of the linearly stacked ones. Such learning circumstances provide an opportunity for further exploration in the potential of methods for feature selection before including all features in the intermediate features. The evaluation of the data is task with respect to the models is shown in Table \ref{tab:eval_pan2020}.

\begin{table}[H]
 \centering
         \caption{Comparison of representations on the \textit{PAN2020} data set without background knowledge (LM) with models incorporating text knowledge graph embeddings (KG). LR in the representation column denotes the linear regression learner and SGD denotes the StochasticGradientDescent learner.}\vspace{0.3cm}
        \resizebox{\textwidth}{!}{\begin{tabular}{|l|c|c|c|c|c|}
            \hline
            Representation & Accuracy & F1 - score & Precision & Recall  \\ \hline
            LR(LM) &  0.6200 & 0.6481 & 0.6034 & 0.7000 \\  \hline
            LR(KG) &   0.6750  & 0.6859 & 0.6635 & 0.7100 \\  \hline
            LR(LM + KG) &   0.6200  & 0.6481 & 0.6034 &0.7000 \\  \hline \hline
            SGD(LSA + TransE + RotatE) & 0.7200 & \textbf{0.7348} & \textbf{0.6900}  &
            \textbf{0.7900} \\ \hline \hline
            SOTA (literature)  \cite{buda:2020} & \textbf{0.7500} & x & x & x \\ \hline

          \end{tabular}}
        \label{tab:eval_pan2020}
\end{table}

\subsection{Task 4: COVID-19} The text based representation of the model outperformed the derived KG representation in terms of all of the metrics. However, the combined representation of the text and knowledge present, significantly improved the score, with the biggest gain from the joint-intermediate representations.  The best-performing representation for this task was the one that was learned on the concatenated representation via SNN with 1024 nodes. This data set did not contain metadata information, so we ommited the KG-ENITTY evaluation. The evaluation of the data is task with respect to the models is shown in Table \ref{tab:eval_covid19}.

\begin{table}[H]
 \centering
        \caption{Comparison of representations on the \textit{COVID-19} data set without background knowledge (LM) with models incorporating text knowledge graph embeddings (KG). LR in the representation column denotes the linear regression learner and SNN denotes the Shallow Neural Network learner.}\vspace{0.3cm}
        \begin{tabular}{|l|c|c|c|c|c|}
            \hline
            Representation & Accuracy & F1 - score & Precision & Recall  \\ \hline
            LR(LM) &  0.9285 & 0.9320 & 0.9275 & 0.9366 \\  \hline
            LR(KG) &  0.8379 & 0.8422 & 0.8582 & 0.8268 \\  \hline
            LR(LM+KG) &  0.9369 & 0.9401 & 0.9347 & 0.9455 \\  \hline \hline
            SNN(LM+KG) & \textbf{0.9570} & 0.9569 & \textbf{0.9533} & \textbf{0.9652} \\ \hline \hline
            SOTA (literature) \cite{glazkova2020g2tmn} & x & \textbf{0.9869} & x & x \\ \hline
          \end{tabular} 
        \label{tab:eval_covid19}
\end{table}

The proposed method of stacking ensembles of representations outscored all other representations for all of the problems. The gain in recall and precision is evident for every problem, since the introduction of conceptual knowledge informs the textual representations about the concepts and the context. The best-performing models were the ones that utilized the textual representations and the factual knowledge of concepts appearing in the data.

\section{Qualitative results}
\label{sec:results-ql}

In the following section we further explore the constructed multi-representation space. In Subsection \ref{subsec:subspaces}, we are interested in whether it is possible to pinpoint which parts of the space were the most relevant for a given problem. In Subsection \ref{sec:word_features}, we analyze whether predictions can be explained with the state-of-the-art explanation methods.

\subsection{Relevant feature subspaces}
\label{subsec:subspaces}
\begin{figure}[!h]
    \centering
    \includegraphics[width=1 \linewidth]{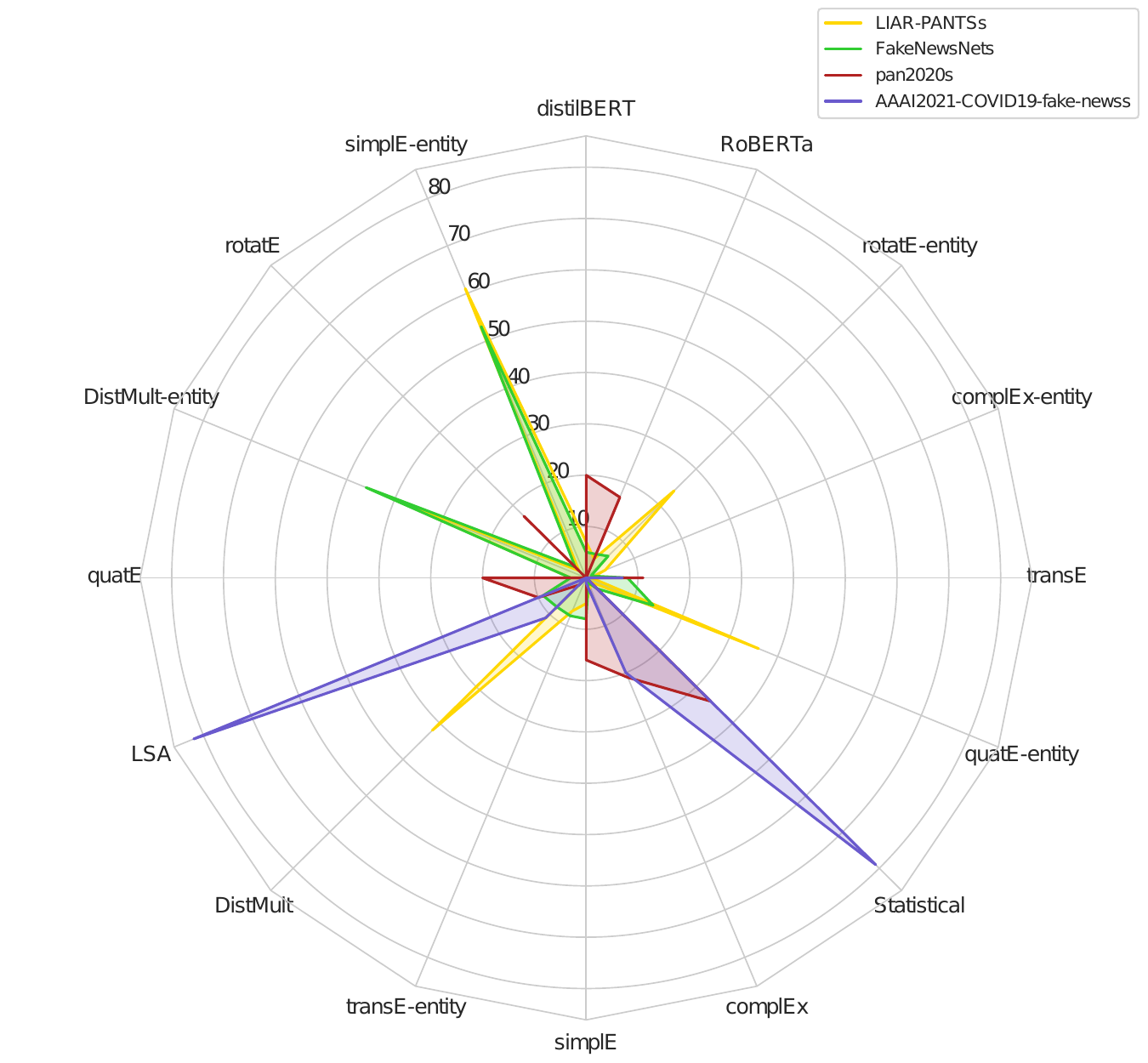}
    \vspace{-1cm}
    \caption{Overview of the most relevant feature subspaces for individual data sets.}
    \label{fig:radial-relevance}
\end{figure}

We next present a procedure and the results for identifying the key feature subspaces, relevant for a given classification task. We extract such features via the use of \emph{supervised feature ranking}, i.e. the process of prioritizing individual features with respect to a given target space. In this work we considered mutual information-based ranking~\cite{kraskov2011erratum}, as the considered spaces were very high dimensional (in both dimensions).
As individual features are mostly latent, and as such non-interpretable, we are interested in what proportion the top $k$ features correspond to a given subspace (e.g., the proportion of BoW features). In this way, we assessed the relevance of a given feature subspace amongst the top features. For the purpose of investigating such subspace counts across different data sets, we present the radial plot-based visualization, shown in Figure~\ref{fig:radial-relevance}.
The radial plot represents the global top ranked feature subspaces. It can be observed that very different types of features correspond to different data sets. For example, the LSA- and statistics-based features were the most relevant for the AAAI data set, however irrelevant for the others. On the other hand, where the knowledge graph-based type of features was relevant, we can observe that multiple different KG-based representations are present. A possible explanation for such behavior is that, as shown in Table~\ref{tab:kg_rel}, methods are to some extent complementary with respect to their expressive power, and could hence capture similar patterns.
\begin{figure*}[!h]
\centering
\subfigure[FakeNewsNet]{
    \includegraphics[width=0.48\textwidth]{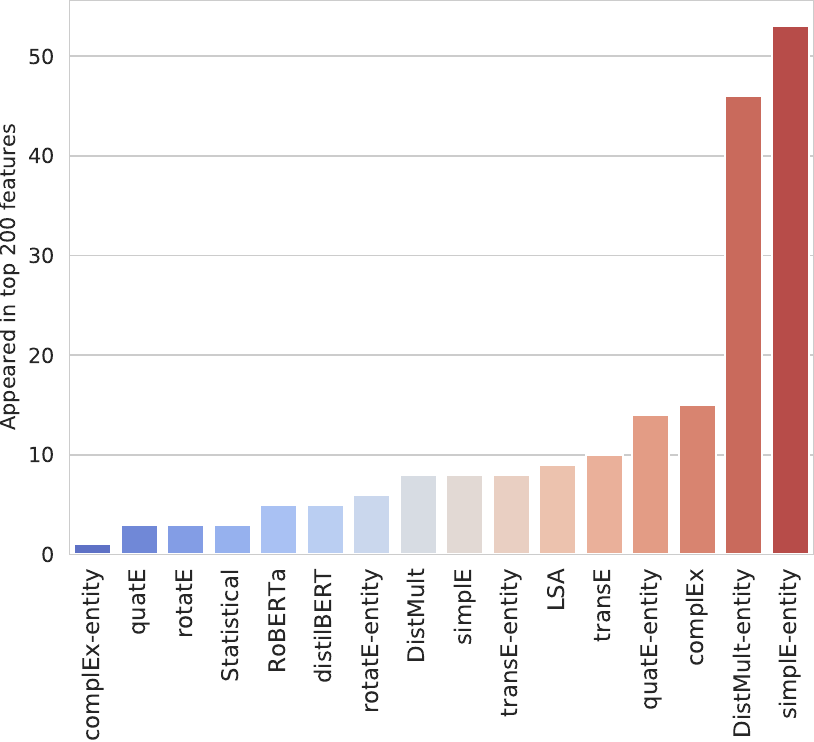}}
\subfigure[LIAR-PANTS]{
    \includegraphics[width=0.48\textwidth]{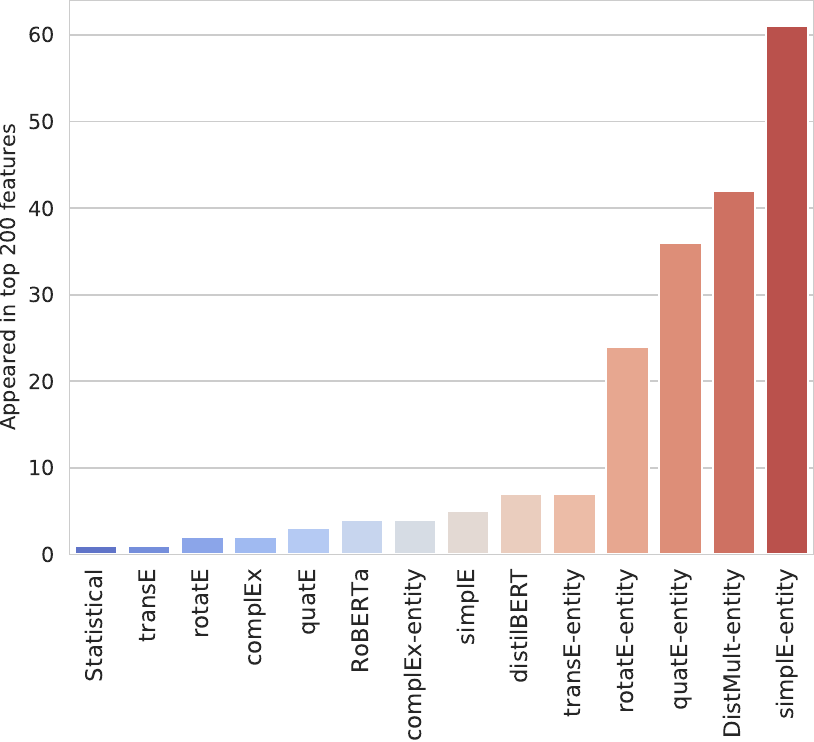}}
    \subfigure[AAAI-COVID19]{
    \includegraphics[width=0.48\textwidth]{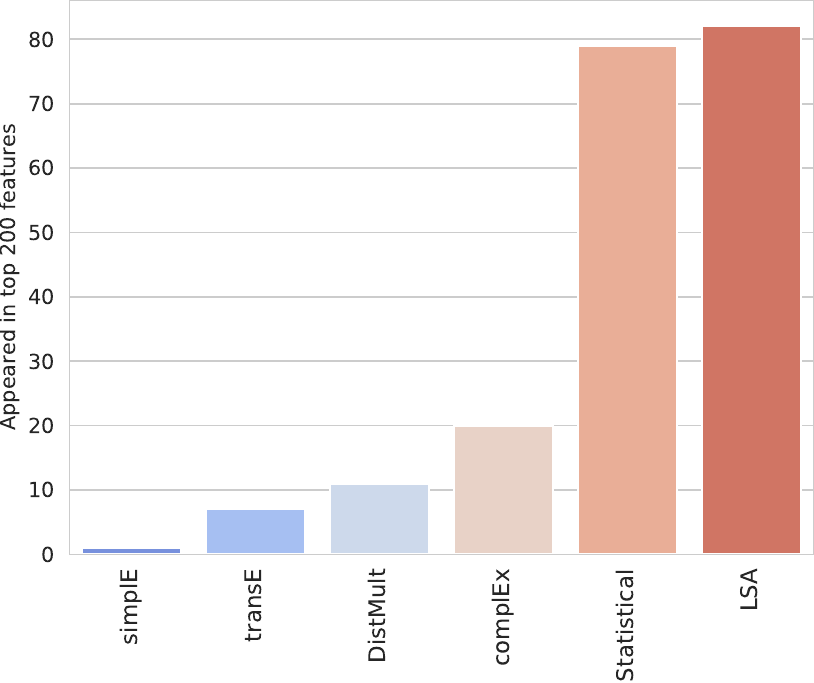}}
\subfigure[PAN2020]{
    \includegraphics[width=0.48\textwidth]{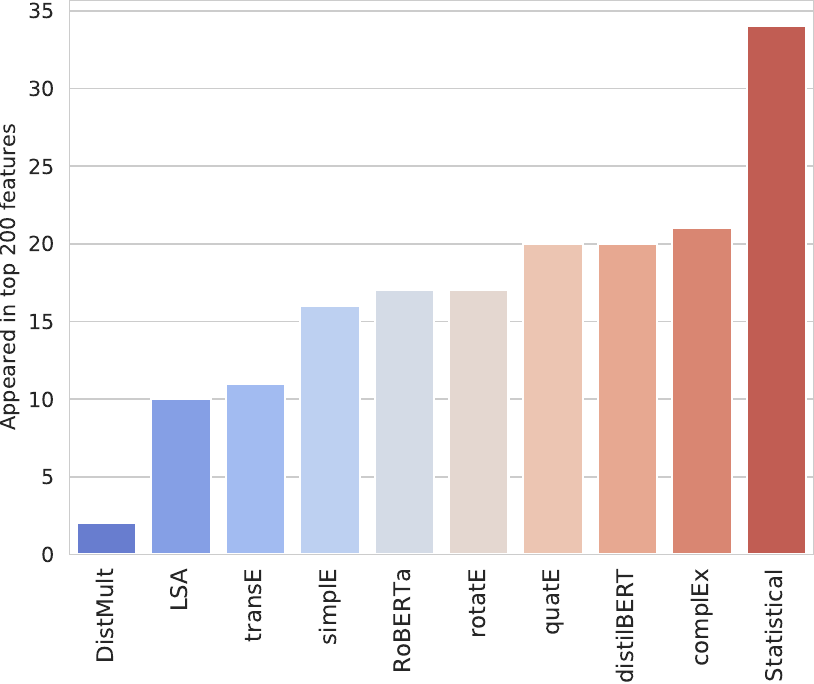}}
\caption{Inspection of ranked subspaces for individual data sets. Note that not all feature types are present amongst the top 200 features according to the feature ranking, indicating that for data sets like AAAI-COVID19, e.g., mostly LSA and statistical features are sufficient.}
    \label{fig:ranking-individual}
\end{figure*}
Individual data sets are inspected in Figure ~\ref{fig:ranking-individual}. For different data sets, different subspaces were the most relevant. For example, for the \emph{FakeNewsNet}, the DistMult and simplE-based representations of given entities were the most frequently observed types of features in top 200 features. This parameter was selected with the aim to capture only the top-ranked features -- out of thousands of features, we hypothesize that amongst the top 200 key subspaces are represented.
The simplE-based features were also the most relevant for the \emph{LIAR-PANTS} data set. However, for the \emph{AAAI-COVID19} data set, the statistical and LSA-based features were the most relevant. A similar situation can be observed for the \emph{PAN2020} data set, where statistical features were the most relevant. The observed differences in ranks demonstrate the utility of multiple representations and their different relevance for individual classification tasks. By understanding the dominating features, one can detect general properties of individual data sets; e.g., high scores of statistical features indicate punctuation-level features could have played a prominent role in the classification. On the contrary, the dominance of entity embeddings indicates that semantic features are of higher relevance.
Note that to our knowledge, this study is one of the first to propose the radial plot-based ranking counts as a method for global exploration of the relevance of individual feature subspaces.

\subsection{Exploratory data analysis study on the knowledge graph features from documents}

\begin{figure*}[h!t]
\centering
\subfigure[FakeNewsNet]{
    \includegraphics[width=0.48\textwidth]{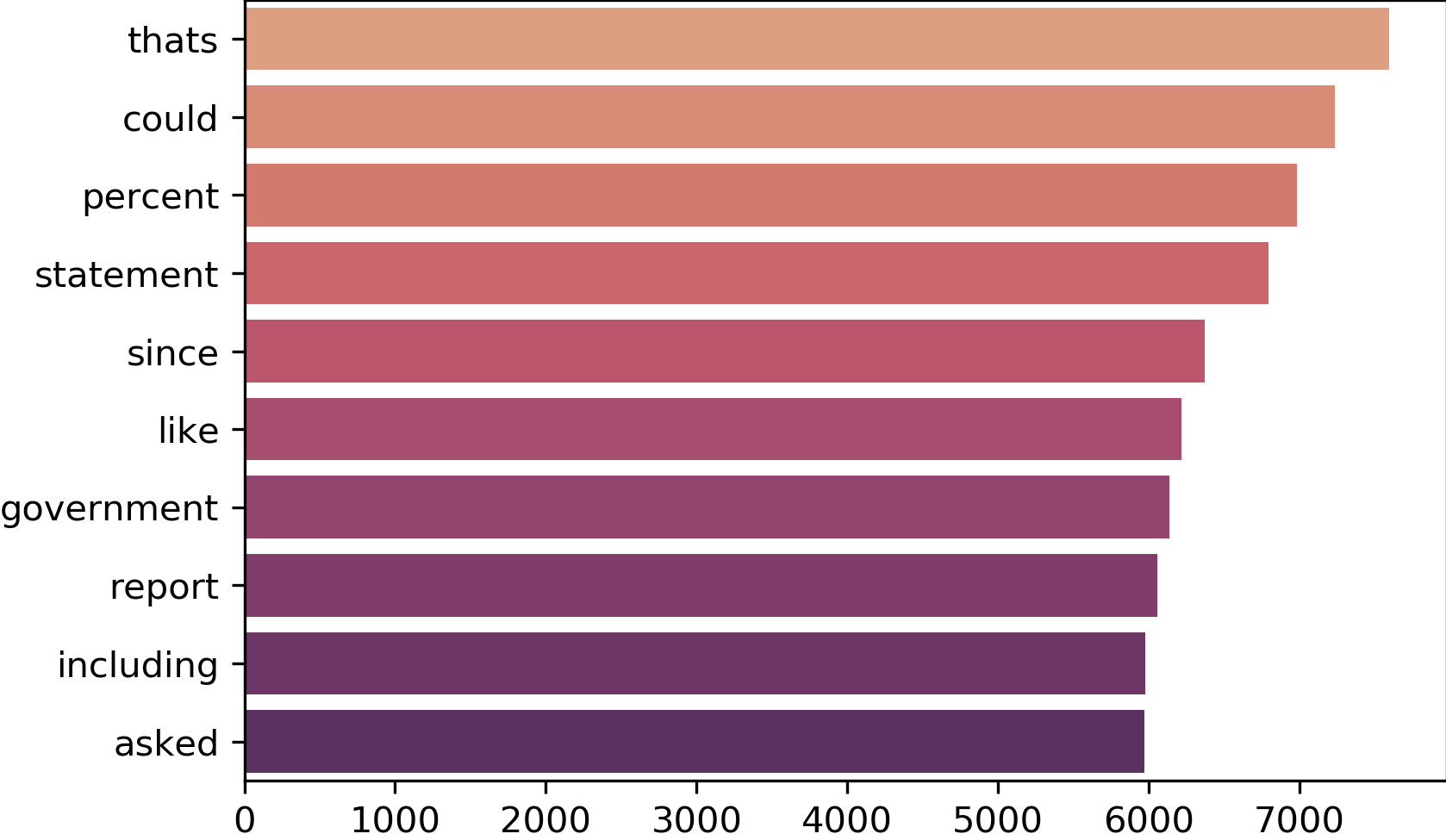}}
\subfigure[FakeNewsNet-ENTITY]{
    \includegraphics[width=0.48\textwidth]{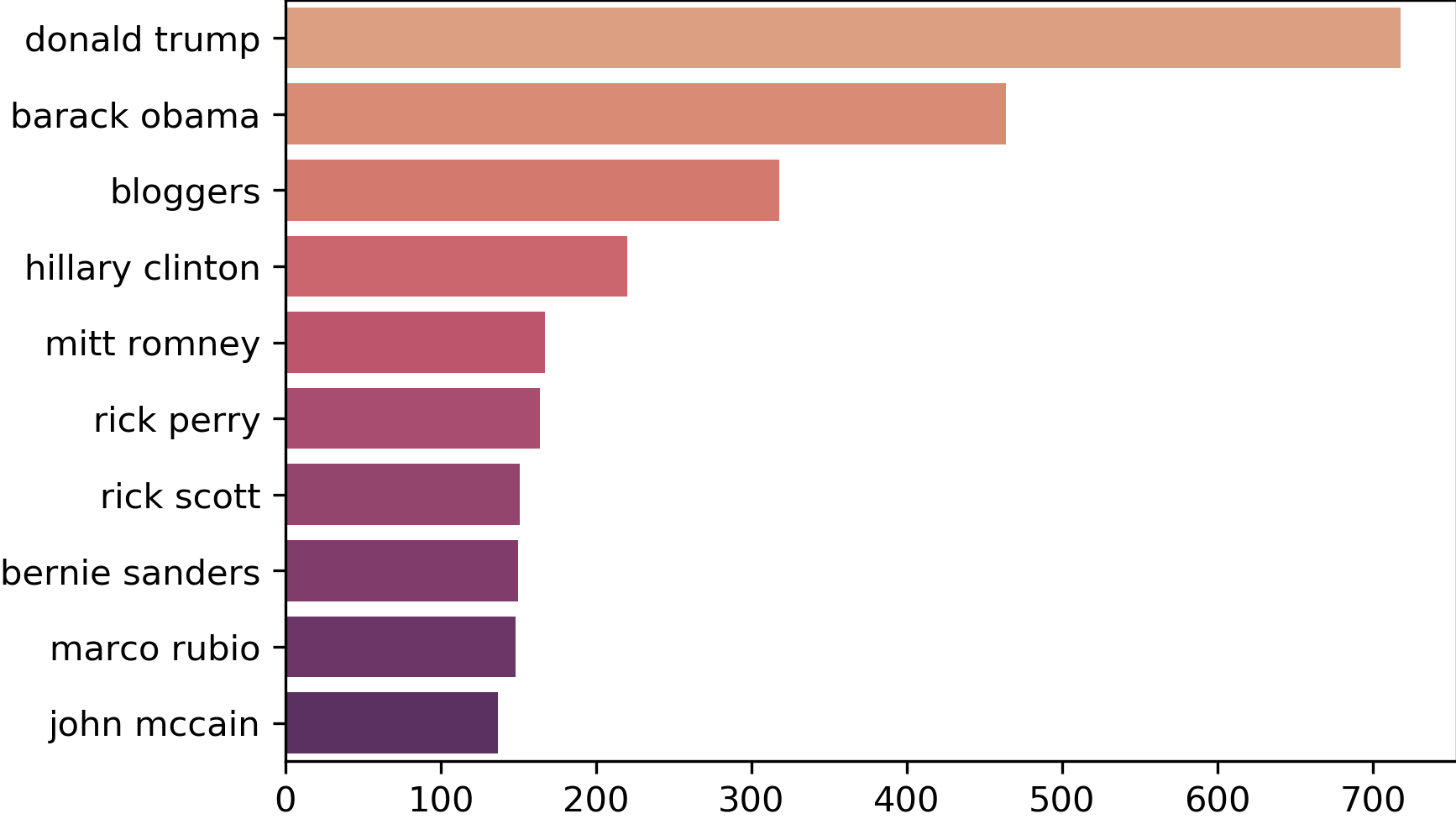}}
\subfigure[LIAR-PANTS]{
    \includegraphics[width=0.48\textwidth]{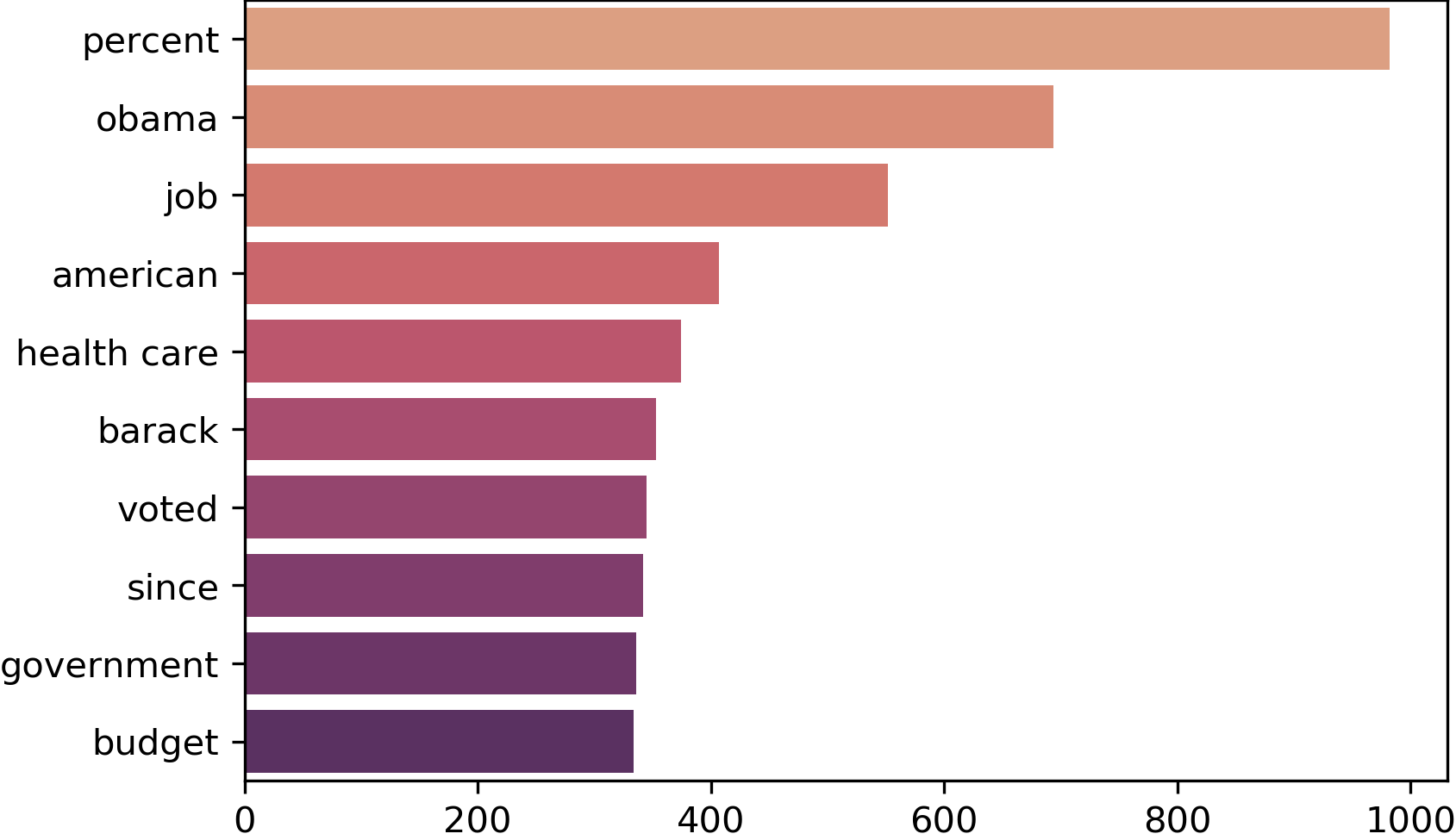}}
\subfigure[LIAR-PANTS-ENTITY]{
    \includegraphics[width=0.48\textwidth]{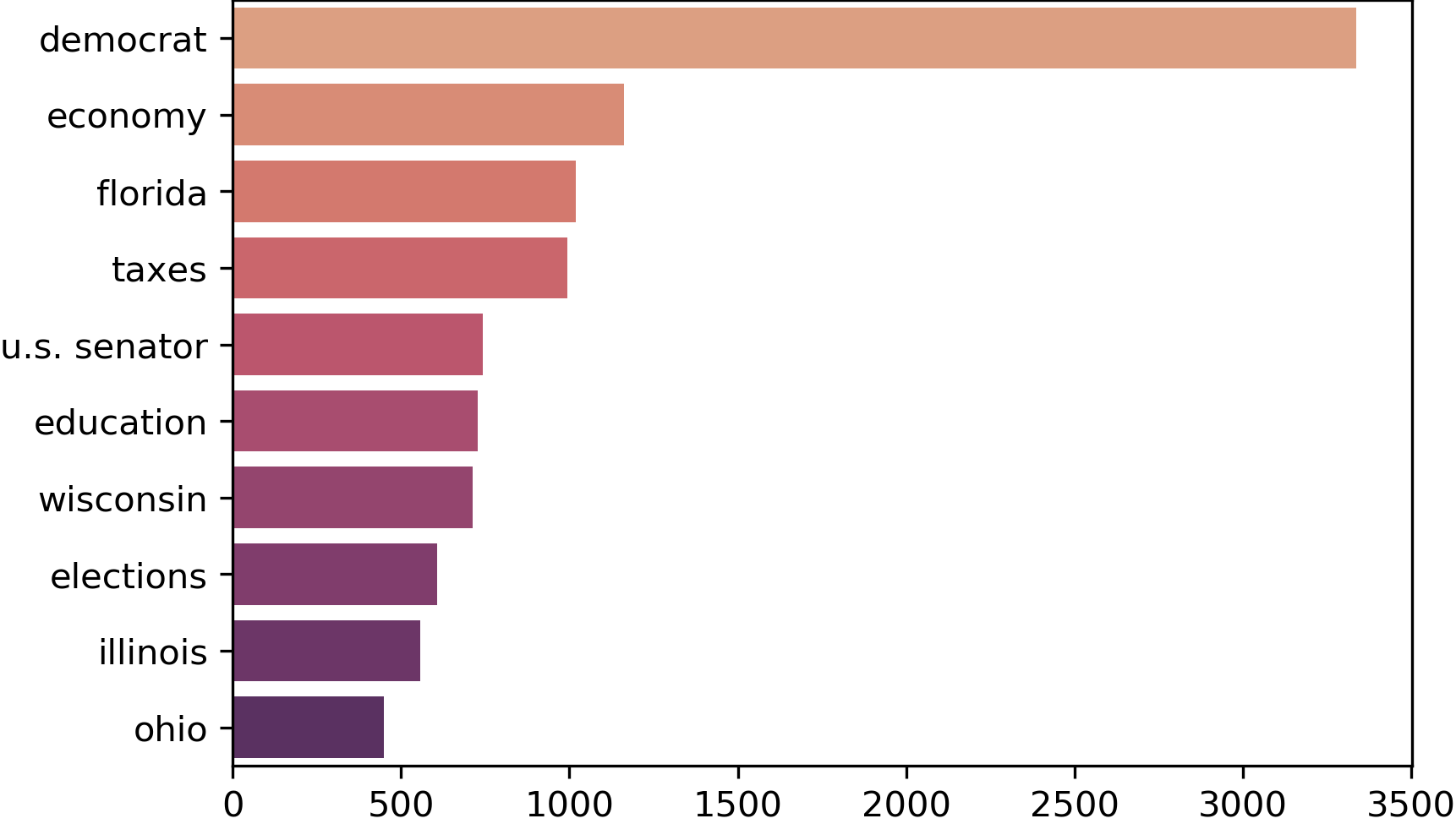}}
\subfigure[PAN2020]{
    \includegraphics[width=0.45\textwidth]{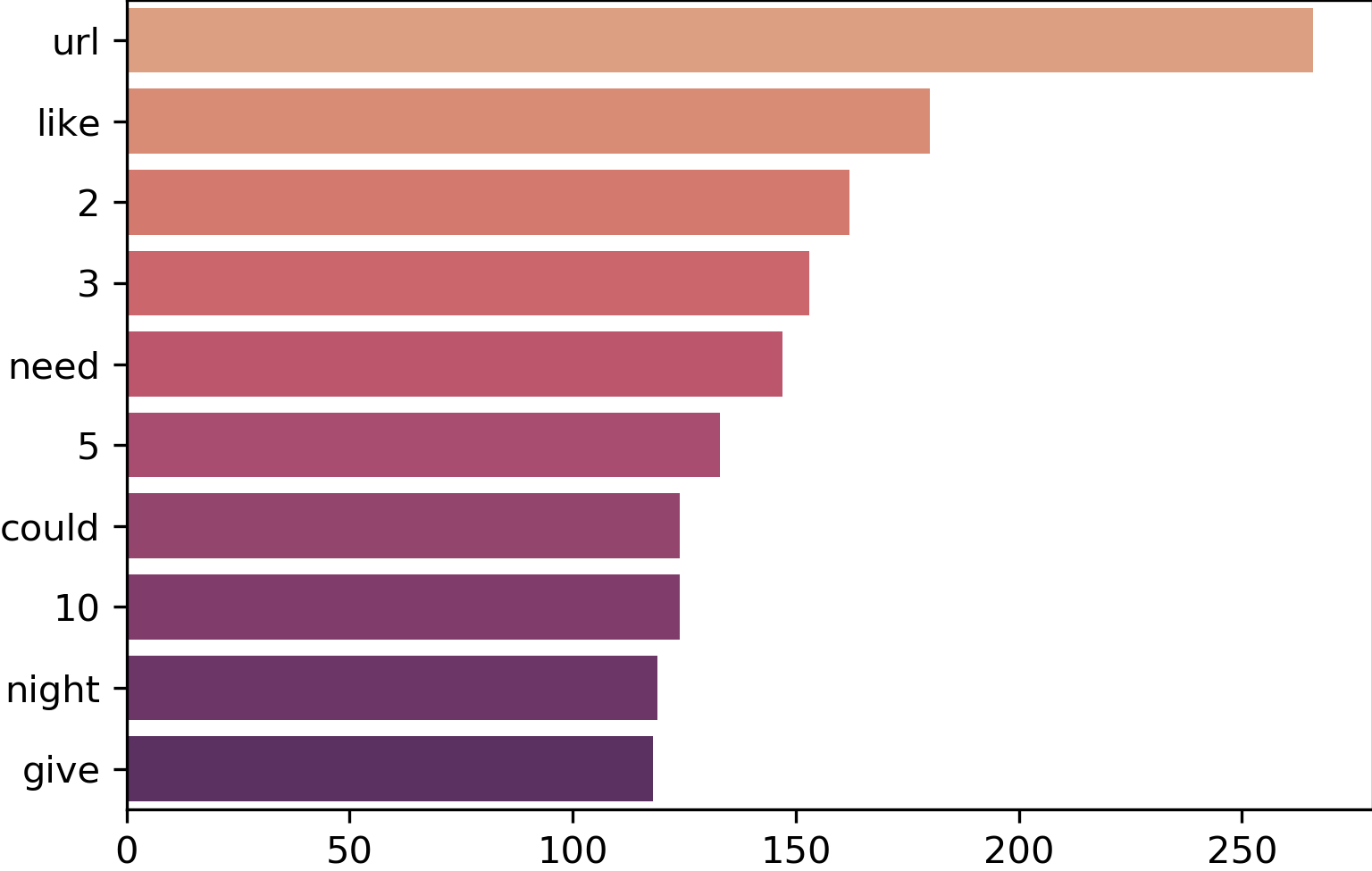}}
\subfigure[AAAI-COVID19]{
    \includegraphics[width=0.48\textwidth]{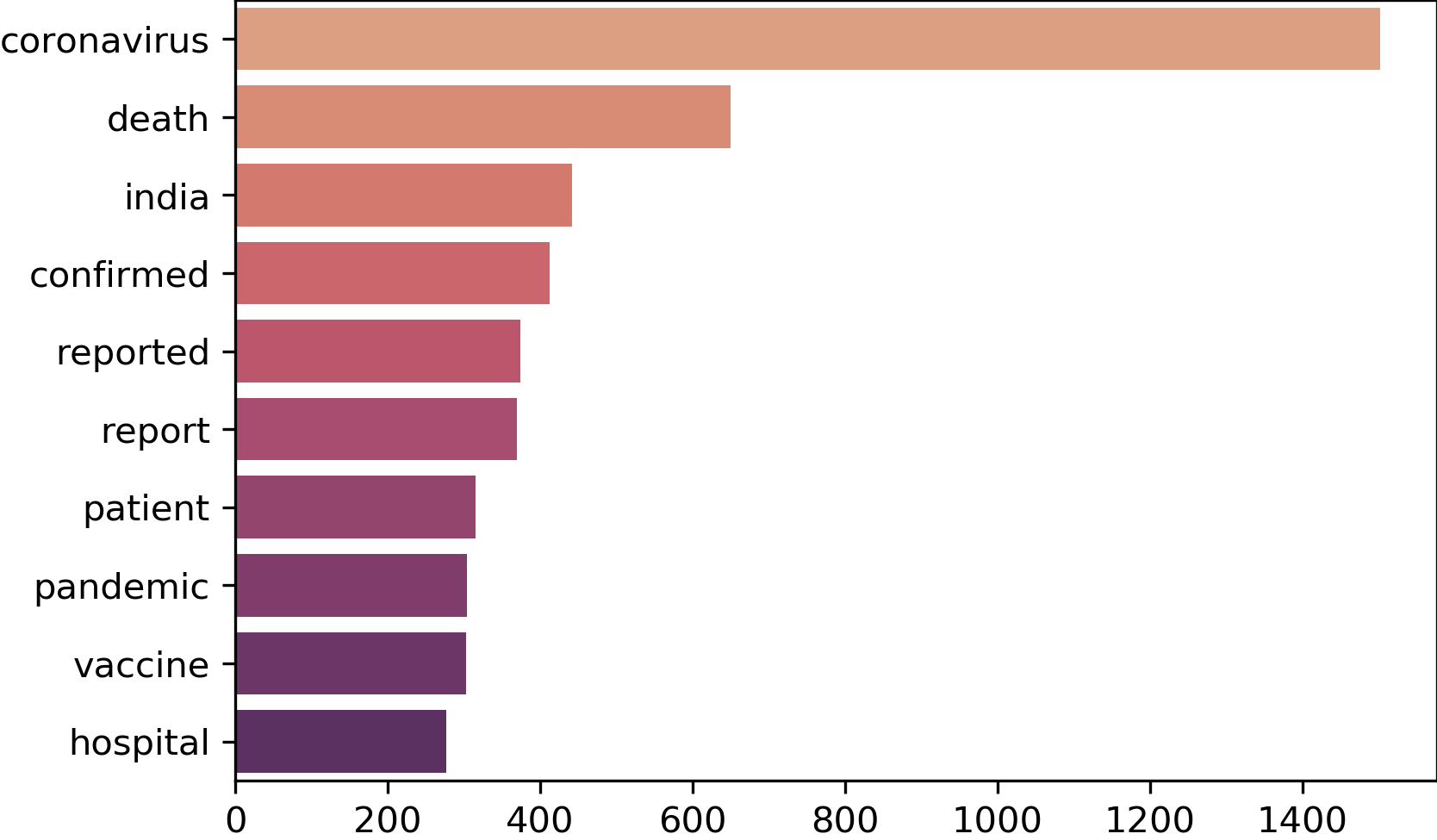}}
\caption{Most common concepts from the WikiData5m KG per article (training data) of the data sets. For the \textit{FakeNewsNet} and \textit{LIAR} data sets, we additionally report the most popular present concepts from the metadata. The x-axis reports the number of occurrences, while the y-axis reports the given concept. }
    \label{fig:train_concepts}
\end{figure*}

In this section we analyze how representative the concept matching is. As described in Subsection \ref{subsec:kg_representations} for each document we first generate the n-grams and extract those present in the KG. For each data set we present the top 10 most frequent concepts that were extracted. First we analyze the induced concepts for all four data sets, followed by the concepts derived from the document metadata for the \textit{LIAR} and \textit{FakeNewsNet} dataset.  The retrieved concepts are shown in Figure \ref{fig:train_concepts}.

The data sets that focus on fake news in the political spectrum (\textit{LIAR} and \textit{FakeNewsNet}) appear to be described by concepts such as \textit{government} and \textit{governmental institutions}, as well political topics revolving around \textit{budget} and \textit{healthcare}. In the case of the metadata representation \textit{Donald Trump} and \textit{Barack Obama} appear as most common. From the general metadata the political affiliation \textit{democrat} comes out on top, followed by political topics such as \textit{economy, taxes, elections} and \textit{education}.
Concepts related to the \textit{coronavirus} such as \textit{death, confirmed and reported cases, patients, pandemic, vaccine, hospital} appeared as the most representative in the \textit{COVID-19} data set. Twitter posts are of limited length and of very versatile nature, making the most common concept in the \textit{PAN2020} data set \textit{URLs} to other sources. Following this, numbers and verbs describing the state of the author such as \textit{need, give, could, and like}. Examples of tweets with present words are given in ~\ref{appendix-real-fake}.

We finally discuss the different concepts that were identified as the most present across the data sets. Even though in data sets like FakeNewsNet and LIAR-PANTS, the most common concepts include well-defined entities such as e.g., 'job', the PAN2020 mapping indicates that this is not necessarily always the case. Given that only for this data set most frequent concepts also include e.g., numbers, we can link this observation to the type of the data -- noisy, short tweets. Having observed no significant performance decreases in this case, we conducted no additional denoising ablations, even though such endeavor could be favourable in general.

\par Next we analyze how much coverage of concepts per data set has the method acquired. We present the distribution of induced knowledge graph concepts per document for every data set in the Appendix in Figure \ref{fig:distb_concepts}. The number of found concepts is comparable across data sets. 

The chosen data sets have more than $98\%$ of their instances covered by additional information, from one or more concepts. For the \textit{LIAR} data set we fail to retrieve concepts only for $1.45\%$ of the instances, for \textit{COVID-19} only for $0.03\%$ instances. In the case of \textit{PAN2020} and \textit{LIAR} data sets we succeed to provide one or more concepts for all examples. Additional distribution details are given in ~\ref{appendix-distribution}.
\subsection{Evaluation of word features in the data}
\label{sec:word_features}

To better understand data sets and obtained models, we inspected words in the \emph{COVID-19 Fake News detection} set as features of the prediction model. We were interested in words that appeared in examples with different contexts which belonged to the same class. To find such words, we evaluated them with the TF-IDF measure, calculated the variance of these features separately for each class and extracted those with the highest variance in their class.  

\begin{figure}[H]
    \centering
    \includegraphics[width=\linewidth]{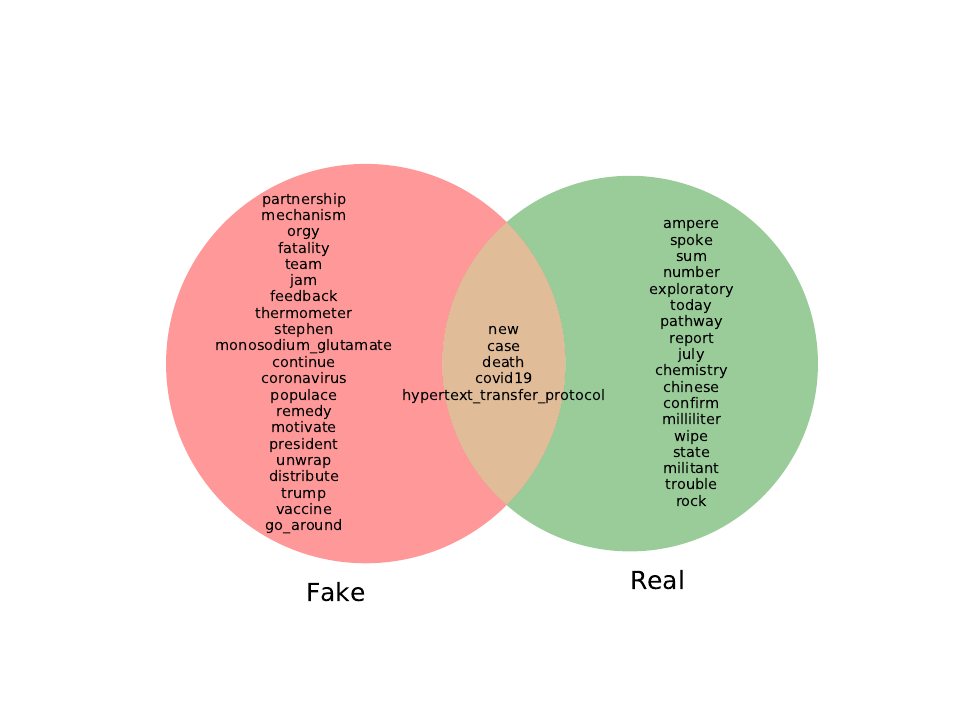}
    \vspace{-2cm}
    \caption{Words with the highest variance in their class. This is the first step towards providing understandable explanations of what affects the classification.}
    \label{fig:variance}
\end{figure}

\begin{figure}[H]
    \centering
    \includegraphics[width=\linewidth]{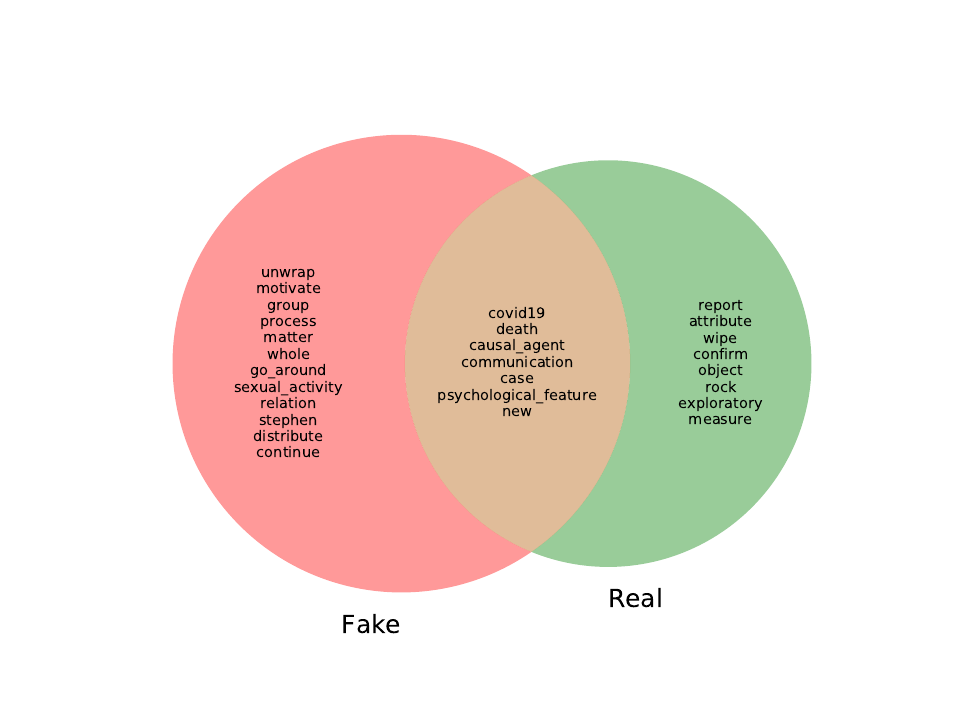}
    \vspace{-1.5cm}
    \caption{We used ReEx with Wordnet to generalize words with the highest variance in their class, and produce understandable explanations.}
    \label{fig:generalized}
\end{figure}

We mapped the extracted words to WordNet \cite{wn} and generalized them using Reasoning with Explanations (ReEx) \cite{9378668} to discover their hypernyms, which can serve as human  understandable explanations. 
Figure \ref{fig:variance} shows words with the highest variance in their respective class, while Figure \ref{fig:generalized} shows found hypernyms of words with the highest variance for each of the classes.

If examined separately, most words found based on variance offer very little as explanations. A couple of words stand out, however; since this is a COVID news data set it is not surprising that words such as "new", "covid19", "death" and "case" are present across different news examples in both classes. Because COVID-19 related news and tweets from different people often contain contradictory information and statements, there must be fake news about vaccines and some substances among them, which could explain their inclusion among words appearing in examples belonging to the "fake" class. Words found in examples belonging to the "real" class seem to be more scientific and concerning measurements, for example, "ampere", "number", "milliliter".


After generalizing words found with variance we can examine what those words have in common. "Causal agent" is a result of the generalization of words in both fake and real classes, which implies that news of both classes try to connect causes to certain events. These explanations also reveal that different measures, attributes and reports can be found in examples belonging to the "real" class.

\section{Discussion}
\label{sec:discussion}
The fake news problem space captured in the aforementioned data sets showed that no single representation or an ensemble of representation works consistently for all problems -- different representation ensembles improve performance for different problems. For instance the author profiling - PAN2020 problem gained performance increase from only a subset of representations the \textit{TransE} and \textit{SimplE} KG derived concepts. As for the FakeNewsNet, the best-performing model was a heterogeneous ensemble of all of the constructed representations and the metadata representations. 
\par The evaluation of the proposed method also showed that the KG only representations were good enough in the case of \textit{PAN2020, LIAR and COVID-19}, where they outperformed the text-only based representations. This represents a potential of researching models based both on contextual and factual knowledge while learning the language model. Wang et al. \cite{wang-etal-2014-knowledge} reported that such approaches can introduce significant improvement; with the increase of the newer methods and mechanisms popular in NLP today we believe this is a promising research venue.
\par Different knowledge embedding methods capture different relational properties. For this study we performed a combination with models that covered Symmetry, Anti-symmetry, Inversion, Transitivity and Composition property. The solutions to some problems benefit from some properties while others benefit from others, in order to explore the possibility one can perform a search through the space of combinations of the available KG models. However exhaustive search can introduce significant increase in the memory and time complexity of learning models. One way to cope with this problem is to apply some regularization to the learner model which would learn on the whole space. The goal of this would be to omit the insignificant combinations of features to affect the predictions of the model. Another approach would be to perform feature selection and afterwards learn only on the representations that appear in the top $k$ representative features.
\section{Conclusions}
\label{sec:conclusions}
    We compared different representations methods for text, graphs and concepts, and proposed a novel method for merging them into a more efficient representation for detection of fake news. We analysed statistical features, matrix factorization embedding LSA, and neural sentence representations sentence-bert, XLM, dBERT, and RoBERTa. We proposed a concept enrichment method for document representations based on data from the WikiData5m knowledge graph. The proposed representations significantly improve the model expressiveness and improve classification performance in all tackled tasks.
    \par The drawbacks of the proposed method include the memory consumption and the growth of the computational complexity with the introduction of high dimensional spaces. In order to cope with this scalability we propose exploring  some dimensionality-reduction approaches such as UMAP {\cite{mcinnes2018umap}} that map the original space to a low-dimensional manifold. Another problem of the method is choosing the right approach for concept extraction from a given text.  Furthermore, a potential drawback of the proposed method is relatively restrictive entity-to-document mapping. By adopting some form of fuzzy matching, we believe we could as further work further improve the mapping quality and with it the resulting representations.
    \par For further work we propose exploring attention based mechanisms to derive explanations for the feature significance of a classification of an instance. Additionally we would like to explore how the other aggregation methods such as the \textit{AGG-TF} and the \textit{AGG-TF-IDF} perform on the given problems. The intensive amount of research focused on the Graph Neural Networks represents another potential field for exploring our method.  The combination of different KG embedding approaches captures different patterns in the knowledge graphs.
 
 The code is freely accessible at ~\url{https://gitlab.com/boshko.koloski/codename_fn_b/, https://github.com/bkolosk1/KBNR}.

\section*{Acknowledgments}
 This paper is supported by European Union’s Horizon 2020 research and innovation programme under grant agreement No. 825153, project EMBEDDIA (Cross-Lingual Embeddings for Less-Represented Languages in European News Media). The authors also acknowledge the financial support from the Slovenian Research Agency for research core funding for the programme Knowledge Technologies (No. P2-0103 and P6-0411), the project CANDAS (No. J6-2581), the CRP project V3-2033 as well as the young researcher's grant of the last author. 
\bibliography{mybibfile}

\appendix

        

\section{Examples of real and fake tweets}
\label{appendix-real-fake}
In this section we present some examples of real and fake tweets with words present (bold). 

\subsection{Real}
\begin{itemize}
\item \textbf{fatality},\#IndiaFightsCorona: India’s Total Recoveries continue to rise cross 32.5 lakh today 5 Statffes contribute 60\% of total cases 62\% of active cases and 70\% of total fatality reported in India \#StaySafe \#IndiaWillWin https://t.co/KRn3GOaBNp

\item \textbf{team},An important part of our work is data collection and analysis At 11:30pm every day our data Team collates results received from all testing laboratories to inform Nigerians of the number of new \#COVID19 cases Results not received at this time are reported the next day https://t.co/Nyo6NeImRk

\item \textbf{partnership},Finally we launched the first real version of the COVID Racial Data Tracker in partnership with @DrIbram and the @AntiracismCtr. This has been a major effort by our project's volunteers—and we hope it will be useful to communities across the country. https://t.co/hTyV0MA5tA
team,In @followlasg our rapid response team is working with NFELTP to strengthen community testing for \#COVID19 in LGAs. The team provides support to newly reactivated LGA walk-in testing sites for increased testing capacity access and awareness of \#COVID19 at the grassroot level. https://t.co/MnIu3OBT3v

\item \textbf{fatality},\#IndiaFightsCorona Health Ministry reviews COVID Management \&amp; Response in 15 districts across 5 States exhibiting high caseload and fatality.
\end{itemize}
\subsection{Fake}
\begin{itemize}
\item \textbf{state},India has lost over 50000 individuals to coronavirus till date. In view of the rising coronavirus cases Bihar government extends lockdown in the state till 6 September. At Nationalist Congress Party chief Sharad Pawar’s residence four people tested positive for \#coronavirus. \url{https://t.co/LqGJHHVr2g}
\item \textbf{report},Leaked Report Says There Are \'Too Many Humans\' On The Planet \url{https://t.co/03kvl3oOXU} \#globalwarming \#coronavirus \#conspiracy
\item \textbf{today},"???Covid is never going away! This is the beach today in Raleigh, North Carolina.??"
\item \textbf{report},"In an Aaj Tak news report the Chinese prime minister said ""Reading Quran and offering namaz is the only cure for COVID-19."""
\item \textbf{chinese},"In an Aaj Tak news report the Chinese prime minister said ""Reading Quran and offering namaz is the only cure for COVID-19."""
\end{itemize}

\section{Distribution of concepts}
\label{appendix-distribution}
\begin{figure}[H]
    \centering
    \caption{Distribution of concepts extracted from the WikiData5m KG per article in the data sets.}
    \begin{tabular}{c}
    \subfigure[FakeNewsNet]{\includegraphics[width =0.90\textwidth]{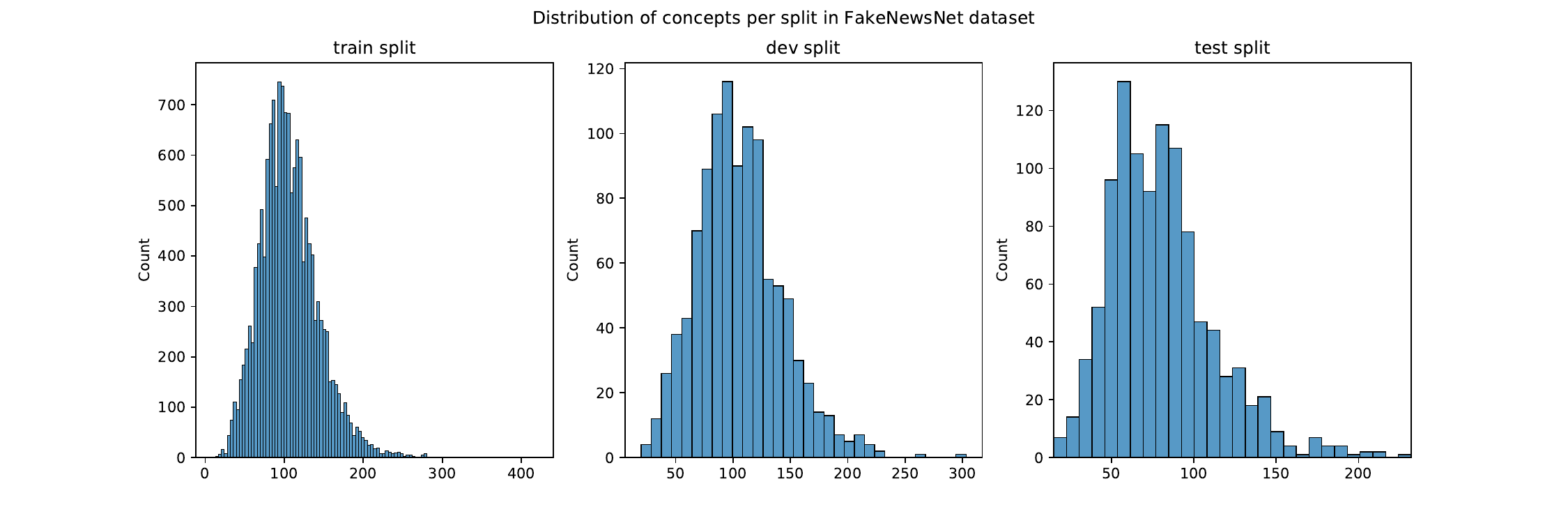}}  \\
    \subfigure[LIAR-PANTS]{\includegraphics[width =0.90\textwidth]{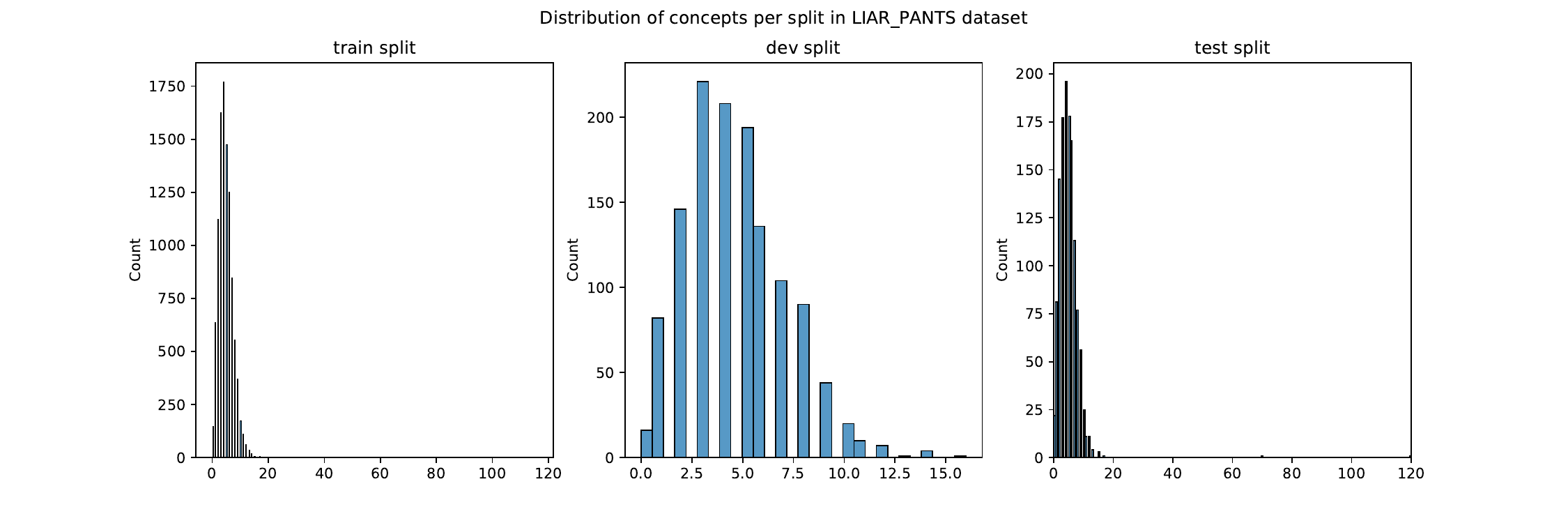}} \\
    \subfigure[AAAI-COVID19]{\includegraphics[width
    =0.90\textwidth]{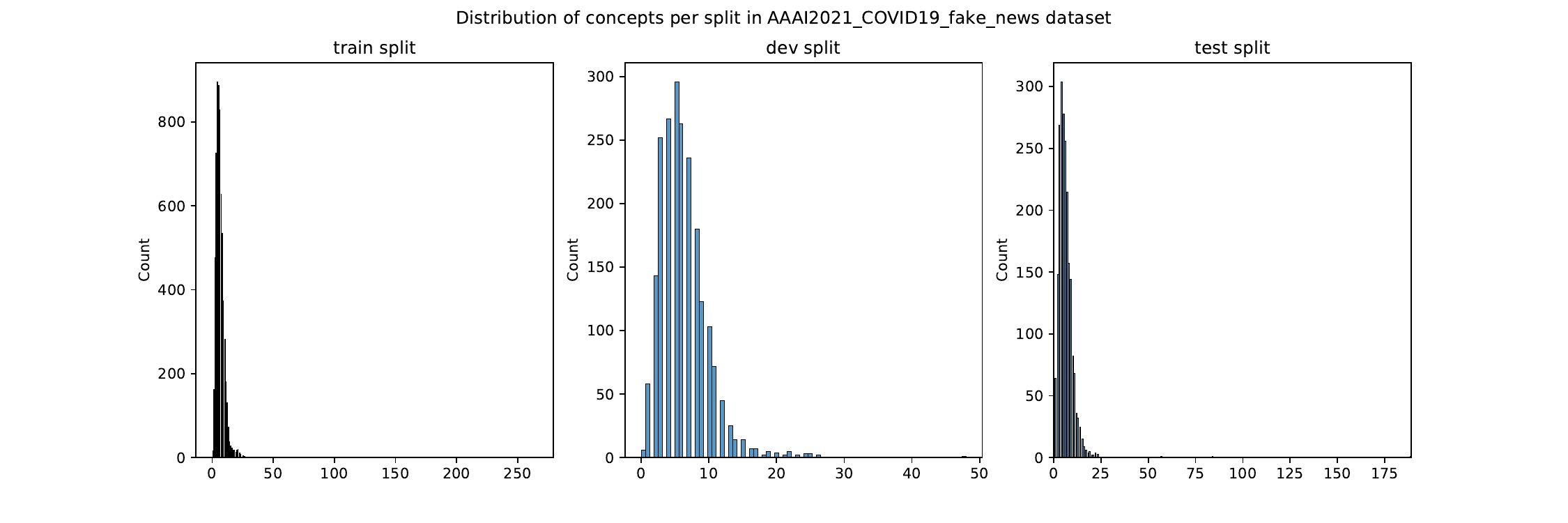}}  \\
    \subfigure[PAN2020]{\includegraphics[width =0.90\textwidth]{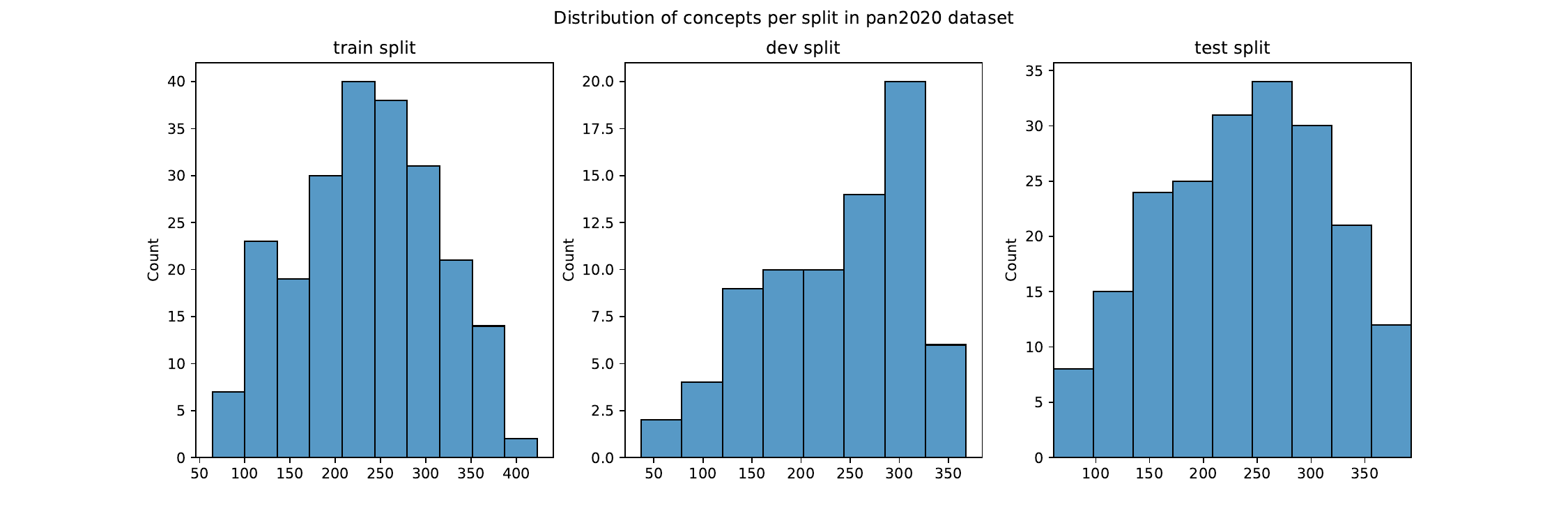}} \\
    \end{tabular}
    \label{fig:distb_concepts}
\end{figure}

In this subsection we showcase the distribution of concepts per each data set, shown in Figure \ref{fig:distb_concepts}.

\section{Performance of individual feature spaces}
We report the performances of individual representations presented as a part of this work next.
\subsection{Evaluation of all subsets of spaces}
In this subsection we explore how combining various spaces affect the performance. Due to the high-cardinality of the document and knowledge-graph embedding we sample $10\%$ with respect to the distribution of lables as in the original distribution. The only exception is the PAN2020 dataset where we use the whole dataset, due to the small number of examples. For every problem we evaluate all the possible combinations consisted of $KG$ representatiosn and $LM$ represetntations, in all-in-all $11$ representations making evaluated in total $2^{11}-1=2047$ combinations of features, on which we learn LogisticRegression classifier with various values of regularization $C \in \{1,0.1,0.01,0.001\}$. For every problem we showcase the best 10 and the worst 10 combinations of features, evaluated at four different score techniques.

\subsubsection{LIAR}
\begin{figure}[H]
    \centering
    \resizebox{\textwidth}{!}{\includegraphics{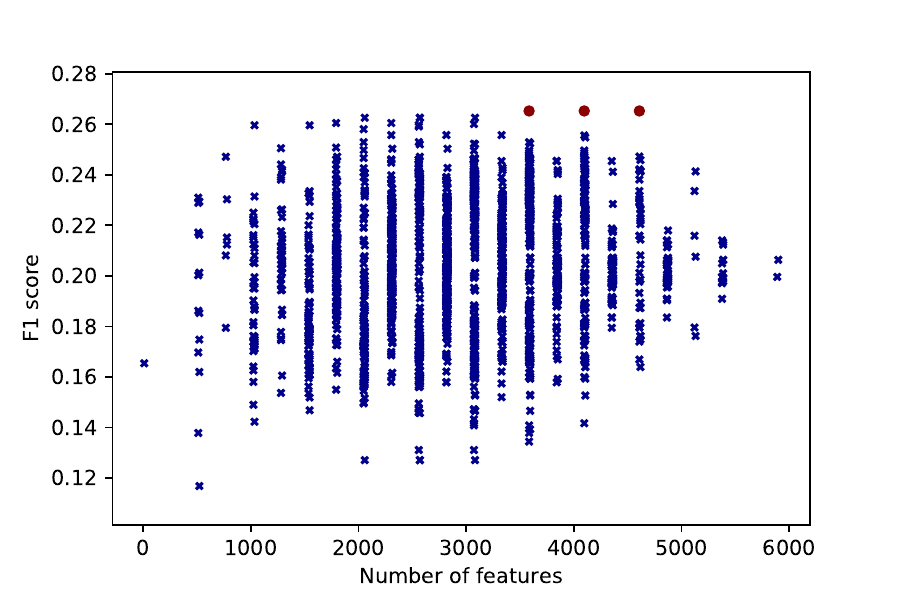}}
    \caption{The interaction of dimensions and the F1-score for the LIAR problem. The red dots represent the highest scoring models.}
    \label{fig:liar_f1_dims}
\end{figure}

The representations that captured only statistical and lexical features show low importance to the task when combined, resulting in an F1-score of 11.68\%. The additional combination of lexical and contextual spaces provided improvement to the scores. The most significant gain on performance concerning the f1-score came with the combination of the QuatE and the simplE knowledge graph features with the dBERT model, improving the score by 11.42\%. Multiple representations landed among the highest F1-score of 26.53\%, the most interesting one is that the combination of DistilBERT and XLM model with statistical features and rotatE knowledge graph embedding yielded top performance. The dependence of the number of features and the f1-scores is represented in Figure \ref{fig:liar_f1_dims}. The worst-performing combinations are listed in Table \ref{tab:liar_worst}, while the best-performing combinations are listed in Table \ref{tab:liar_best}.
\begin{table}[H]
    \centering
\resizebox{\textwidth}{!}{
\begin{tabular}{lrrrrr}
\toprule
combination &  dimensions &  f1\_scores &  accuracy\_score &  precision\_score &  recall\_score \\
\midrule
LSA\_stat &         522 &   0.116782 &        0.141732 &         0.117917 &      0.121464 \\
rotate\_roBERTa\_stat\_XLM &        2058 &   0.127043 &        0.149606 &         0.127742 &      0.129400 \\
rotate\_LSA\_roBERTa\_stat\_XLM &        2570 &   0.127043 &        0.149606 &         0.127742 &      0.129400 \\
transe\_rotate\_roBERTa\_stat\_XLM &        2570 &   0.127043 &        0.149606 &         0.127742 &      0.129400 \\
transe\_rotate\_LSA\_roBERTa\_stat\_XLM &        3082 &   0.127043 &        0.149606 &         0.127742 &      0.129400 \\
transe\_rotate\_quate\_distmult\_simple\_LSA &        3072 &   0.131043 &        0.149606 &         0.137023 &      0.130886 \\
rotate\_quate\_distmult\_simple\_LSA &        2560 &   0.131043 &        0.149606 &         0.137023 &      0.130886 \\
complex\_rotate\_quate\_LSA\_roBERTa\_XLM &        3584 &   0.134385 &        0.141732 &         0.139119 &      0.134308 \\
LSA &         512 &   0.137799 &        0.165354 &         0.138862 &      0.142240 \\
complex\_transe\_rotate\_quate\_distmult\_simple\_LSA &        3584 &   0.137810 &        0.157480 &         0.143607 &      0.137337 \\
\bottomrule
\end{tabular}
}
    \caption{Liar worst 10 representation combinations.}
    \label{tab:liar_worst}
    
\end{table}

\begin{table}[H]
    \centering
\resizebox{\textwidth}{!}{\begin{tabular}{lrrrrr}
\toprule
combination &  dimensions &  f1\_scores &  accuracy\_score &  precision\_score &  recall\_score \\
\midrule
transe\_rotate\_DistilBERT\_LSA\_XLM &        3072 &   0.260089 &        0.275591 &         0.260826 &      0.261883 \\
quate\_simple\_DistilBERT &        1792 &   0.260485 &        0.275591 &         0.277576 &      0.257641 \\
transe\_quate\_simple\_DistilBERT &        2304 &   0.260485 &        0.275591 &         0.277576 &      0.257641 \\
rotate\_DistilBERT\_stat\_XLM &        2058 &   0.262555 &        0.275591 &         0.266784 &      0.262160 \\
rotate\_DistilBERT\_LSA\_stat\_XLM &        2570 &   0.262555 &        0.275591 &         0.266784 &      0.262160 \\
transe\_rotate\_DistilBERT\_LSA\_stat\_XLM &        3082 &   0.262555 &        0.275591 &         0.266784 &      0.262160 \\
transe\_rotate\_DistilBERT\_stat\_XLM &        2570 &   0.262555 &        0.275591 &         0.266784 &      0.262160 \\
complex\_transe\_quate\_distmult\_simple\_DistilBERT\_LSA\_roBERTa &        4608 &   0.265255 &        0.283465 &         0.269992 &      0.263042 \\
complex\_quate\_distmult\_simple\_DistilBERT\_roBERTa &        3584 &   0.265255 &        0.283465 &         0.269992 &      0.263042 \\
complex\_transe\_quate\_distmult\_simple\_DistilBERT\_roBERTa &        4096 &   0.265255 &        0.283465 &         0.269992 &      0.263042 \\
\bottomrule
\end{tabular}
}
    \caption{LIAR best 10 representation combinations.}
    \label{tab:liar_best}
\end{table}

\subsubsection{FakeNewsNet}
\begin{figure}[H]
    \centering
    \resizebox{\textwidth}{!}{\includegraphics{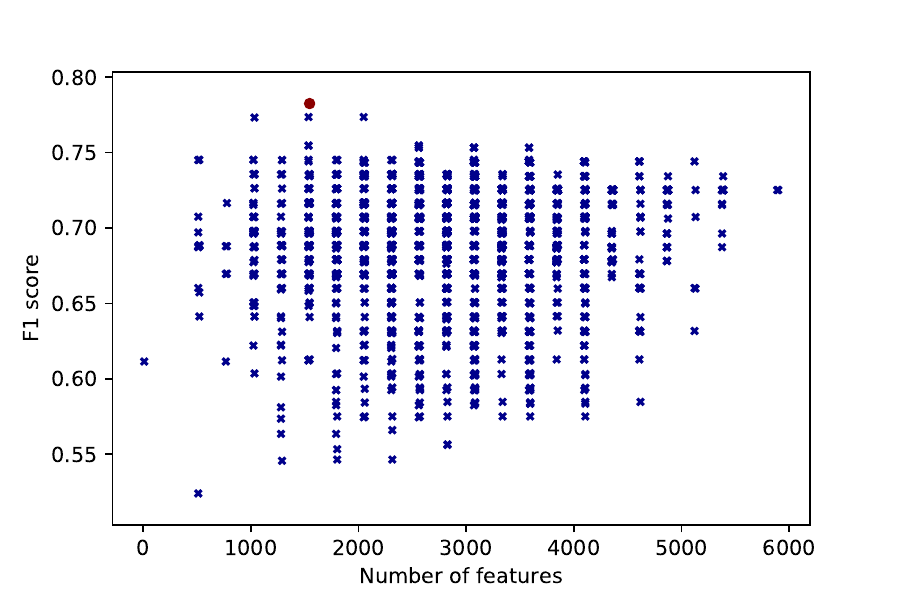}}
    \caption{The interaction of dimensions and the F1-score for the FakeNewsNet problem. The red dots represent the highest scoring models.}
    \label{fig:fnn_f1_dims}
\end{figure}

Knowledge graph and their combinations generated too general spaces that scored lowest on the dataset. The lowest scoring representation is the one based only on the TransE KG embedding method. Notable improvement was seen with introduction of the contextual representation. The best performing model for this problem was the one that combined features from knowledge graphs that preserve various relations(the ComplEx, TransE, and RotatE embeddings) and the simple stylometric representation. The dependence of the number of features and the f1-scores is represented in Figure \ref{fig:fnn_f1_dims}.  The worst-performing combinations are listed in Table \ref{tab:fnn_worst}, while the best-performing combinations are listed in Table \ref{tab:fnn_best}.
\begin{table}[H]
    \centering
    \resizebox{\textwidth}{!}{\begin{tabular}{lrrrrr}
\toprule
combination &  dimensions &  f1\_scores &  accuracy\_score &  precision\_score &  recall\_score \\
\midrule
transe &         512 &   0.524066 &        0.528302 &         0.582348 &      0.572545 \\
rotate\_stat\_XLM &        1290 &   0.545714 &        0.547170 &         0.557471 &      0.559524 \\
rotate\_LSA\_stat\_XLM &        1802 &   0.546524 &        0.547170 &         0.561957 &      0.563616 \\
transe\_rotate\_LSA\_stat\_XLM &        2314 &   0.546524 &        0.547170 &         0.561957 &      0.563616 \\
transe\_rotate\_stat\_XLM &        1802 &   0.553384 &        0.556604 &         0.560606 &      0.563244 \\
transe\_rotate\_quate\_LSA\_stat\_XLM &        2826 &   0.556248 &        0.556604 &         0.573953 &      0.575521 \\
transe\_rotate\_quate\_distmult\_stat\_XLM &        2826 &   0.556564 &        0.556604 &         0.584428 &      0.583705 \\
rotate\_XLM &        1280 &   0.563552 &        0.566038 &         0.572143 &      0.575149 \\
transe\_distmult\_XLM &        1792 &   0.563552 &        0.566038 &         0.572143 &      0.575149 \\
rotate\_quate\_distmult\_stat\_XLM &        2314 &   0.566038 &        0.566038 &         0.591518 &      0.591518 \\
\bottomrule
\end{tabular}

}
    \caption{FakeNewsNet worst 10 representation combinations.}
    \label{tab:fnn_worst}
\end{table}

\begin{table}[H]
    \centering
    \resizebox{\textwidth}{!}{\begin{tabular}{lrrrrr}
\toprule
combination &  dimensions &  f1\_scores &  accuracy\_score &  precision\_score &  recall\_score \\
\midrule
complex\_LSA\_roBERTa\_XLM &        2560 &   0.753312 &        0.754717 &         0.761429 &      0.772321 \\
transe\_rotate\_quate\_distmult\_roBERTa\_XLM &        3584 &   0.753312 &        0.754717 &         0.761429 &      0.772321 \\
transe\_rotate\_simple &        1536 &   0.754630 &        0.754717 &         0.780425 &      0.784598 \\
complex\_rotate\_quate &        1536 &   0.754717 &        0.754717 &         0.788690 &      0.788690 \\
complex\_transe\_rotate\_simple\_LSA &        2560 &   0.754717 &        0.754717 &         0.788690 &      0.788690 \\
complex\_rotate\_quate\_simple\_LSA &        2560 &   0.754717 &        0.754717 &         0.788690 &      0.788690 \\
complex\_rotate\_stat &        1034 &   0.773262 &        0.773585 &         0.792391 &      0.800223 \\
complex\_transe\_simple\_LSA &        2048 &   0.773585 &        0.773585 &         0.808408 &      0.808408 \\
complex\_simple\_LSA &        1536 &   0.773585 &        0.773585 &         0.808408 &      0.808408 \\
complex\_transe\_rotate\_stat &        1546 &   0.782535 &        0.783019 &         0.798594 &      0.808036 \\
\bottomrule
\end{tabular}}
    \caption{FakeNewsNet best 10 representation combinations.}
    \label{tab:fnn_best}
\end{table}

\subsubsection{PAN2020}
\begin{figure}[H]
    \centering
    \resizebox{\textwidth}{!}{\includegraphics{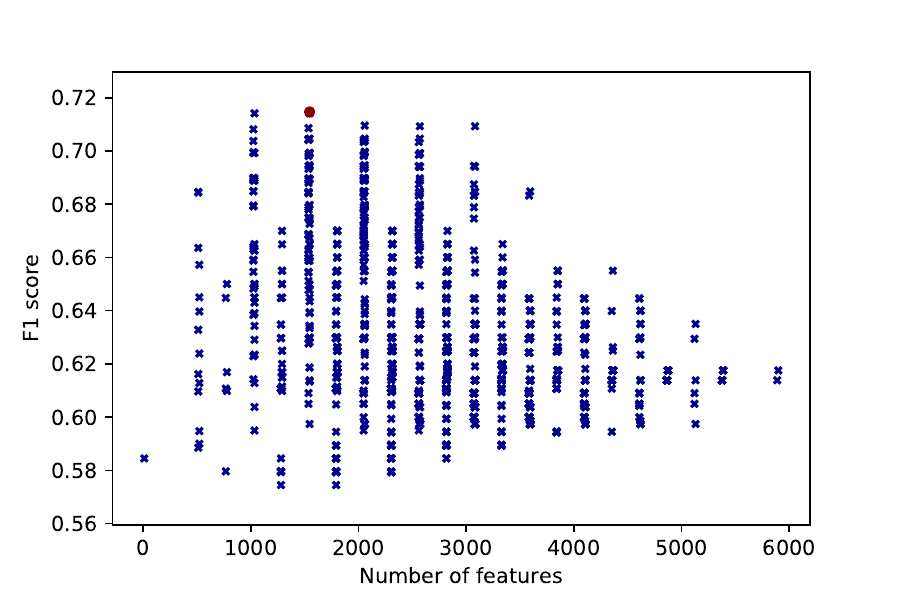}}
    \caption{The interaction of dimensions and the F1-score for the PAN2020 problem. The red dots represent the highest scoring models.}    
    \label{fig:pan20_f1}
\end{figure}

For the PAN2020 problem, the combination of the knowledge graph representations with the contextual-based language representations as XLM ranked the lowest, with a F1-score of 57.45\%. The problem benefited the most from the LSA representation, the additional enrichment of this space with knowledge graph features improved the score by $14.02\%$. The best-performing model based on ComplEx and QuatE KG embeddings and LSA and statsitical language features, with a dimension of 1546. The worst-performing combinations are listed in Table \ref{tab:pan20_worst}, while the best-performing combinations are listed in Table \ref{tab:pan20_best}. The dependence of the number of features and the f1-scores is represented in Figure \ref{fig:pan20_f1}.

\begin{table}[H]
    \centering
    \resizebox{\textwidth}{!}{\begin{tabular}{lrrrrr}
\toprule
combination &  dimensions &  f1\_scores &  accuracy\_score &  precision\_score &  recall\_score \\
\midrule
complex\_transe\_XLM &        1792 &   0.574479 &           0.575 &         0.575369 &         0.575 \\
complex\_XLM &        1280 &   0.574479 &           0.575 &         0.575369 &         0.575 \\
quate\_LSA\_XLM &        1792 &   0.579327 &           0.580 &         0.580515 &         0.580 \\
quate\_distmult\_XLM &        1792 &   0.579327 &           0.580 &         0.580515 &         0.580 \\
transe\_quate\_distmult\_XLM &        2304 &   0.579327 &           0.580 &         0.580515 &         0.580 \\
transe\_quate\_LSA\_XLM &        2304 &   0.579327 &           0.580 &         0.580515 &         0.580 \\
transe\_LSA\_XLM &        1792 &   0.579327 &           0.580 &         0.580515 &         0.580 \\
complex\_transe\_LSA\_XLM &        2304 &   0.579327 &           0.580 &         0.580515 &         0.580 \\
complex\_LSA\_XLM &        1792 &   0.579327 &           0.580 &         0.580515 &         0.580 \\
LSA\_XLM &        1280 &   0.579327 &           0.580 &         0.580515 &         0.580 \\
\bottomrule
\end{tabular}}
    \caption{PAN2020 worst 10 representation combinations.}
    \label{tab:pan20_worst}
\end{table}
\begin{table}[H]
    \centering
    \resizebox{\textwidth}{!}{\begin{tabular}{lrrrrr}
\toprule
combination &  dimensions &  f1\_scores &  accuracy\_score &  precision\_score &  recall\_score \\
\midrule
complex\_transe\_quate\_distmult\_LSA\_stat &        2570 &   0.704638 &           0.705 &         0.706009 &         0.705 \\
complex\_quate\_distmult\_LSA\_stat &        2058 &   0.704638 &           0.705 &         0.706009 &         0.705 \\
distmult\_LSA &        1024 &   0.708132 &           0.710 &         0.715517 &         0.710 \\
transe\_distmult\_LSA &        1536 &   0.708572 &           0.710 &         0.714198 &         0.710 \\
complex\_transe\_quate\_distmult\_simple\_LSA\_stat &        3082 &   0.709273 &           0.710 &         0.712121 &         0.710 \\
complex\_quate\_distmult\_simple\_LSA\_stat &        2570 &   0.709273 &           0.710 &         0.712121 &         0.710 \\
complex\_transe\_quate\_LSA\_stat &        2058 &   0.709535 &           0.710 &         0.711353 &         0.710 \\
transe\_quate\_LSA\_stat &        1546 &   0.714135 &           0.715 &         0.717633 &         0.715 \\
quate\_LSA\_stat &        1034 &   0.714135 &           0.715 &         0.717633 &         0.715 \\
complex\_quate\_LSA\_stat &        1546 &   0.714650 &           0.715 &         0.716059 &         0.715 \\
\bottomrule
\end{tabular}}
    \caption{PAN2020 best 10 representation combinations.}
    \label{tab:pan20_best}
\end{table}

\subsubsection{COVID-19}
\begin{figure}[H]
    \centering
    \resizebox{\textwidth}{!}{\includegraphics{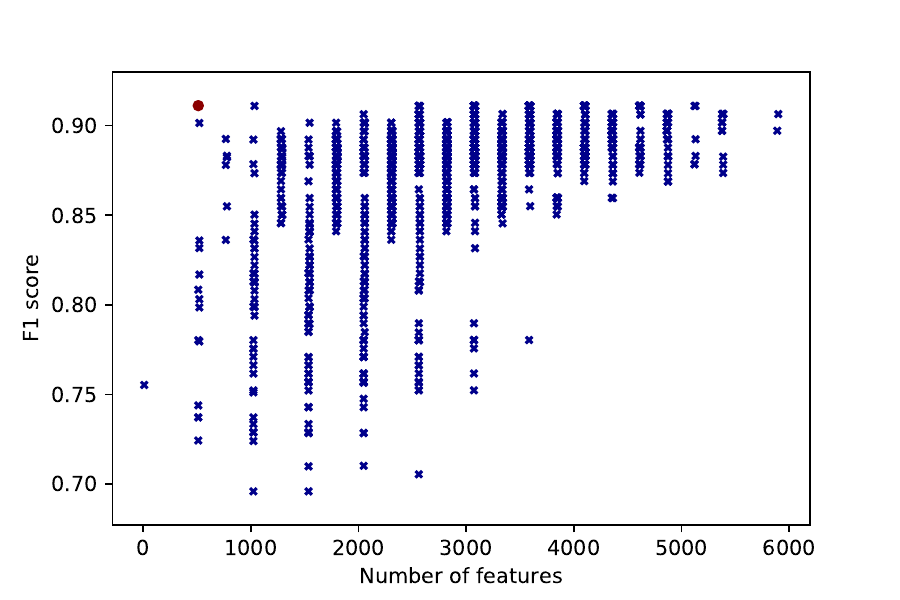}}
    \caption{The interaction of dimensions and the F1-score for the COVID-19 problem. The red dots represent the highest scoring models.}    
    \label{fig:c19_f1_dims}
\end{figure}

Knowledge graph only based representation yielded too general spaces, making for the lowest-performing spaces for the COVID-19 task. Notable improvement for the dataset was achieved by the addition of language models to the knowledge graph representations. The worst-performing combinations are listed in Table \ref{tab:c19_tbls_worst}, while the best-performing combinations are listed in Table \ref{tab:c19_best}. The dependence of the number of features and the f1-scores is represented in Figure \ref{fig:c19_f1_dims}.

\begin{table}[H]
    \centering
\resizebox{\textwidth}{!}{\begin{tabular}{lrrrrr}
\toprule
combination &  dimensions &  f1\_scores &  accuracy\_score &  precision\_score &  recall\_score \\
\midrule
complex\_transe\_distmult &        1536 &   0.695936 &        0.696262 &         0.695893 &      0.696254 \\
complex\_distmult &        1024 &   0.695936 &        0.696262 &         0.695893 &      0.696254 \\
complex\_transe\_rotate\_quate\_distmult &        2560 &   0.705447 &        0.705607 &         0.705607 &      0.706057 \\
transe\_rotate\_distmult &        1536 &   0.709875 &        0.710280 &         0.709790 &      0.710084 \\
complex\_rotate\_quate\_distmult &        2048 &   0.710179 &        0.710280 &         0.710517 &      0.710959 \\
rotate\_distmult &        1024 &   0.724004 &        0.724299 &         0.723941 &      0.724352 \\
complex &         512 &   0.724293 &        0.724299 &         0.725488 &      0.725665 \\
complex\_quate\_distmult &        1536 &   0.728379 &        0.728972 &         0.728379 &      0.728379 \\
complex\_transe\_quate\_distmult &        2048 &   0.728379 &        0.728972 &         0.728379 &      0.728379 \\
transe\_rotate\_quate\_distmult &        2048 &   0.728593 &        0.728972 &         0.728497 &      0.728817 \\
\bottomrule
\end{tabular}}
    \caption{COVID-19 worst 10 representation combinations.}
    \label{tab:c19_tbls_worst}
\end{table}
\begin{table}[H]
    \centering
\resizebox{\textwidth}{!}{\begin{tabular}{lrrrrr}
\toprule
combination &  dimensions &  f1\_scores &  accuracy\_score &  precision\_score &  recall\_score \\
\midrule
transe\_rotate\_quate\_simple\_DistilBERT\_roBERTa &        3584 &   0.910886 &        0.911215 &         0.911770 &      0.910364 \\
transe\_rotate\_distmult\_simple\_DistilBERT\_roBERTa &        3584 &   0.910886 &        0.911215 &         0.911770 &      0.910364 \\
transe\_quate\_distmult\_simple\_DistilBERT\_roBERTa &        3584 &   0.910886 &        0.911215 &         0.911770 &      0.910364 \\
rotate\_quate\_distmult\_simple\_DistilBERT\_roBERTa &        3584 &   0.910886 &        0.911215 &         0.911770 &      0.910364 \\
rotate\_quate\_distmult\_DistilBERT\_LSA\_roBERTa &        3584 &   0.910886 &        0.911215 &         0.911770 &      0.910364 \\
rotate\_distmult\_simple\_DistilBERT\_LSA\_roBERTa &        3584 &   0.910886 &        0.911215 &         0.911770 &      0.910364 \\
transe\_rotate\_quate\_distmult\_simple\_DistilBERT\_LSA\_roBERTa &        4608 &   0.910886 &        0.911215 &         0.911770 &      0.910364 \\
complex\_transe\_rotate\_quate\_distmult\_DistilBERT\_roBERTa &        4096 &   0.910886 &        0.911215 &         0.911770 &      0.910364 \\
complex\_distmult\_simple\_DistilBERT\_LSA\_roBERTa &        3584 &   0.910886 &        0.911215 &         0.911770 &      0.910364 \\
LSA &         512 &   0.911058 &        0.911215 &         0.910916 &      0.911239 \\
\bottomrule
\end{tabular}
}
    \caption{COVID-19 best 10 representation combinations.}
    \label{tab:c19_best}
\end{table}

\subsection{Conclusion} 
    In this section we discuss the main highlights of the extensive ablation studies targeting the performance of different feature space combinations. The main conclusions are as follows.
    
    In the evaluation of spaces study, we analyzed how combining various spaces before learning common joint spaces impacts performance. We can take two different outputs from the study: 
    
    \begin{enumerate}
        \item knowledge graph-based representations on their own are too general for tasks where the main type of input are short texts. However, including additional statistical and contextual information about such texts has shown to improve the performance. The representations that are capable of capturing different types of relation properties (e.g., symmetry, asymmetry, inversion etc.) in general perform better than the others.
        \item We observed no general rule determining the optimal representation combination. Current results, however, indicate, that transfer learning based on different representation types is a potentially interesting research direction. Furthermore, similarity between the spaces could be further studied at the task level.
    \end{enumerate}

\end{document}